\pdfoutput=1

\documentclass[11pt]{article}

\usepackage{EMNLP2023}

\usepackage{times}
\usepackage{latexsym}

\usepackage[T1]{fontenc}

\usepackage[utf8]{inputenc}

\usepackage{microtype}

\usepackage{inconsolata}

\newcommand{\slic}{\textsc{CoLT5}\xspace}
\newcommand{\longt}{\textsc{LongT5}\xspace}

\definecolor{sliccolor}{rgb}{0, 0.46484375, 0.73046875} 
\definecolor{longtcolor}{rgb}{0.9296875, 0.3984375, 0.46484375} 
\definecolor{darkgrey}{rgb}{0.33203125,0.33203125,0.33203125} 
\definecolor{fewnqcolor}{rgb}{0, 0.46484375, 0.73046875} 
\definecolor{fewtqacolor}{rgb}{0.95, 0.52, 0.0} 

\definecolor{othercolor}{rgb}{0.85, 0.75, 0.85} 
\definecolor{answercolor}{rgb}{1.0, 0.7, 0.28} 
\definecolor{questioncolor}{rgb}{0.21, 0.46, 0.53} 

\newcommand{\modelmark}{square*}
\newcommand{\longtmark}{diamond*}
\newcommand{\othermark}{*}
\newcommand{\answermark}{triangle*}
\newcommand{\questionmark}{square*}

\newcommand*{\minWidth}{35}
\newcommand*{\maxValue}{1}
\newcommand{\makeBar}[4]{%
    \hspace{-6.5pt}
  \tikz[baseline]{
    \node[anchor=base,text width=\minWidth,align=#2,inner sep=0pt,inner xsep=\tabcolsep,outer sep=0pt] (n) {\strut{#1}};
    \begin{pgfonlayer}{background}
        {
       \ifdim #1pt > 0.0pt 
            \edef\color{#3} 
            \pgfmathparse{0.5 + abs(#1/(2*\maxValue))}
            \edef\contents{{\pgfmathresult}}
            \fill[font=\boldmath,color=\color] ($(n.north west)!{0.5}!(n.north east)$) rectangle ($(n.south west)!{{\contents}}!(n.south east)$);
      \else
            \edef\color{#4}
            \pgfmathparse{0.5 - abs(#1/(2*\maxValue))}
            \edef\contents{{\pgfmathresult}}
            \fill[font=\boldmath,color=\color] ($(n.north west)!{{\contents}}!(n.north east)$) rectangle ($(n.south west)!{0.5}!(n.south east)$);
      \fi
        }
    \end{pgfonlayer}
  }
    \hspace{-9.5pt}
}

\newcommand{\correlationBar}[1]{\makeBar{#1}{center}{cyan!25}{red!25}}

\usepackage{times}
\usepackage{latexsym}
\usepackage{xspace}
\usepackage{amsmath}
\usepackage{tabularx}
\usepackage{multirow}
\usepackage{booktabs}

\usepackage{xcolor}
\definecolor{dgreen}{rgb}{0,0.5,0}

\usepackage{tikz}
\pgfdeclarelayer{background}
\pgfdeclarelayer{main}
\pgfsetlayers{background,main}
\usetikzlibrary{calc}
\usepackage{ifthen}

\usepackage{graphicx}
\usepackage{caption}
\usepackage{subcaption}
\usepackage{float}
\usepackage{stfloats}
\usepackage{tikz}
\usepackage{pgfplots}
\usepgfplotslibrary{fillbetween}
\usetikzlibrary{patterns}
\pgfplotsset{compat=1.8}

\newenvironment{customlegend}[1][]{%
    \begingroup
    \csname pgfplots@init@cleared@structures\endcsname
    \pgfplotsset{#1}%
}{%
    \csname pgfplots@createlegend\endcsname
    \endgroup
}%

\def\addlegendimage{\csname pgfplots@addlegendimage\endcsname}

\title{\slic: Faster Long-Range Transformers with Conditional Computation}

\author{
Joshua Ainslie\thanks{\- Author contributions are outlined in Appendix \ref{sec:appendix-contributions}. Correspondence author: jainslie@google.com.},~ Tao Lei,~ Michiel de Jong,~ Santiago Onta\~{n}\'{o}n \\
{\bf Siddhartha Brahma},~ {\bf Yury Zemlyanskiy},~ {\bf David Uthus},~ {\bf Mandy Guo} \\
{\bf James Lee-Thorp},~ {\bf Yi Tay},~ {\bf Yun-Hsuan Sung},~ {\bf Sumit Sanghai}
  \AND
  {\rm \Large Google Research}\\
  }

\begin{document}
\maketitle
\begin{abstract}
Many natural language processing tasks benefit from long inputs, but processing long documents with Transformers is expensive -{}- not only due to quadratic attention complexity but also from applying feedforward and projection layers to every token. However, not all tokens are equally important, especially for longer documents. We propose \slic, a long-input Transformer model that builds on this intuition by employing conditional computation, devoting more resources to important tokens in both feedforward and attention layers. We show that \slic achieves stronger performance than \longt with much faster training and inference, achieving SOTA on the long-input SCROLLS benchmark. Moreover, \slic can effectively and tractably make use of extremely long inputs, showing strong gains up to 64k input length.

\end{abstract}

\section{Introduction}\label{sec:intro}

Many natural language processing tasks, such as summarization~\cite{cohan2018arxiv} or question answering over long documents~\cite{joshi2017triviaqa}, require machine learning models to encode long-form text. Processing long documents with a Transformer model is computationally expensive, both because attention cost scales quadratically with input length and because feedforward and attention projection layers have to be applied to each input token.

Over the past few years, many ``efficient Transformer'' approaches have been proposed that reduce the cost of the attention mechanism over long inputs~\cite{child2019generating, ainslie2020etc, beltagy2020longformer, zaheer2020big, wang2020linformer, tay2021long, guo2022longt5}. However, especially for larger models, the feedforward and projection layers actually make up the majority of the computational burden and can render processing long inputs intractable. 

This paper presents \slic (Conditional LongT5), a new family of models that, building on top of \longt~\cite{guo2022longt5}, enables fast processing of long inputs by combining architecture improvements for both attention and feedforward layers. \slic is based on the intuition that some tokens are more important than others, and we can achieve better quality for lower cost by devoting more computation to important tokens. Moreover, the fraction of important tokens is likely to diminish with document length, allowing for tractable processing of long documents.
\begin{figure}[t!]
    \centering
    \includegraphics[width=0.95\columnwidth]{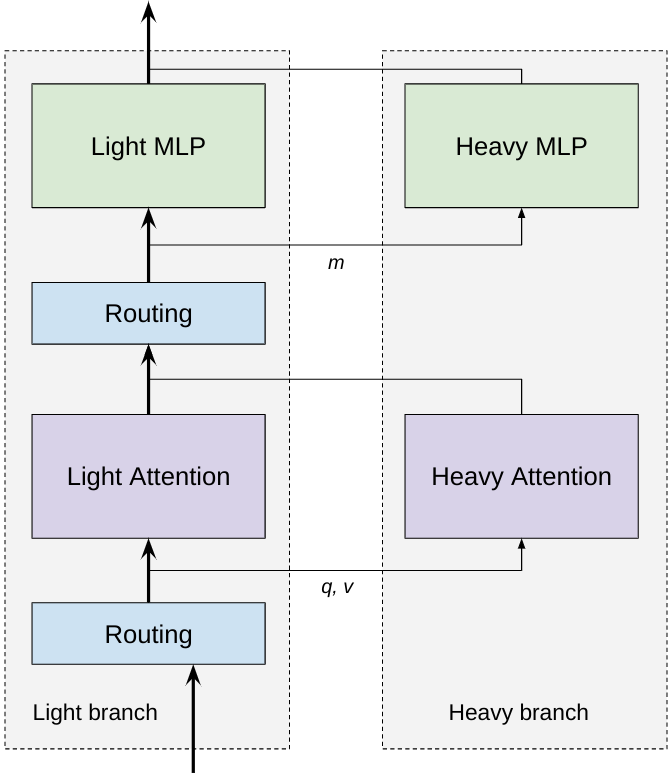}
    \caption{An overview of a \slic Transformer layer with conditional computation. All tokens are processed by light attention and MLP layers, while $q$ routed query tokens perform heavier attention over $v$ routed key-value tokens and $m$ routed tokens are processed by a heavier MLP.}
    \label{fig:slic}
\end{figure}
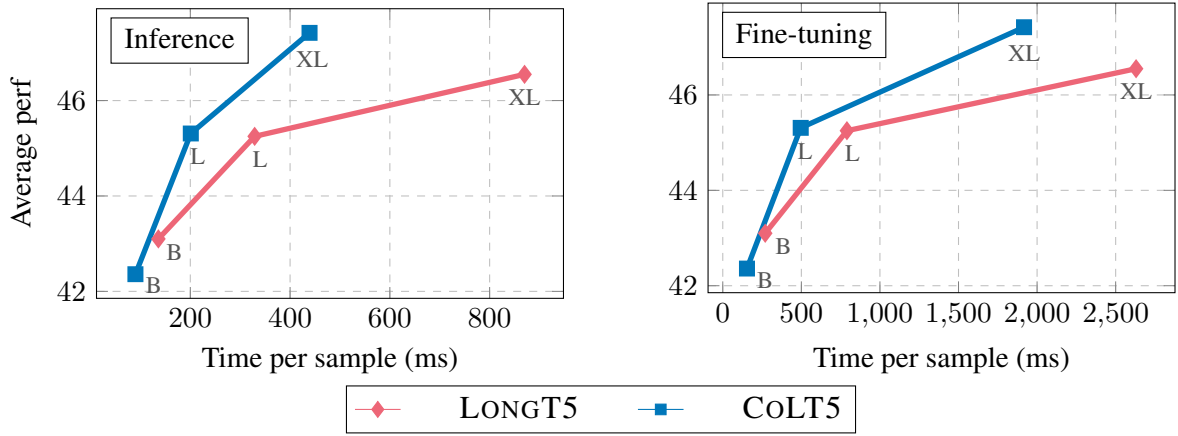
\begin{figure*}[t!]
     \centering
     \begin{subfigure}[t]{0.48\textwidth}
        \centering
        \begin{tikzpicture}[scale=1.0]
            \begin{axis}[
            scale only axis,
            width=0.8\textwidth,
            height=0.5\textwidth,
            xlabel={Time per sample (ms)},
            ylabel={Average perf},
            mark=x,
            x tick label style={log ticks with fixed point},
            ymajorgrids=true,
            xmajorgrids=true,
            xminorticks=true,
            grid style=dashed,
            legend pos={north west},
            ylabel style={align=left, text height=0.2cm},
            ]
            \addlegendimage{empty legend}\addlegendentry{Inference}
            \addplot[color=longtcolor,mark=\longtmark,mark size=2pt, line width=2] table {
                136 43.10
                329 45.25
                870 46.55
            };            
            \addplot[color=sliccolor,mark=\modelmark,mark size=2pt, line width=2] table {
                90 42.36
                201 45.31
                439 47.42
            };
            \node at (axis cs:132.00,43.20) [anchor= north west, color=darkgrey] {\small{B}};
            \node at (axis cs:340.0,45.15) [anchor= north, color=darkgrey] {\small{L}};
            \node at (axis cs:870.0,46.45) [anchor= north, color=darkgrey] {\small{XL}};
            
            \node at (axis cs:90.0,42.50) [anchor= north west, color=darkgrey] {\small{B}};
            \node at (axis cs:215,45.20) [anchor= north, color=darkgrey] {\small{L}};
            \node at (axis cs:444,47.25) [anchor= north, color=darkgrey] {\small{XL}};                 
            \end{axis}
        \end{tikzpicture}          
     \end{subfigure}
     \hfill
     \begin{subfigure}[t]{0.48\textwidth}
        \centering
        \begin{tikzpicture}[scale=1.0]
            \begin{axis}[
            scale only axis,
            width=0.8\textwidth,
            height=0.5\textwidth,
            xlabel={Time per sample (ms)},
            mark=x,
            x tick label style={log ticks with fixed point},
            ymajorgrids=true,
            xmajorgrids=true,
            xminorticks=true,
            grid style=dashed,
            legend pos={north west},
            ylabel style={align=left, text height=0.2cm},
            ]
            \addlegendimage{empty legend}\addlegendentry{Fine-tuning}
            \addplot[color=longtcolor,mark=\longtmark,mark size=2pt, line width=2] table {
                 270 43.10
                790 45.25
                2630 46.55
            };

            \addplot[color=sliccolor,mark=\modelmark,mark size=2pt, line width=2] table {
                154 42.36
                496 45.31
                1917 47.42
            };
            \node at (axis cs:270.00,43.20) [anchor= north west, color=darkgrey] {\small{B}};
            \node at (axis cs:825.0,45.15) [anchor= north, color=darkgrey] {\small{L}};
            \node at (axis cs:2630.0,46.45) [anchor= north, color=darkgrey] {\small{XL}};
            
            \node at (axis cs:154.0,42.50) [anchor= north west, color=darkgrey] {\small{B}};
            \node at (axis cs:525,45.20) [anchor= north, color=darkgrey] {\small{L}};
            \node at (axis cs:1917,47.25) [anchor= north, color=darkgrey] {\small{XL}};          
            \end{axis}
        \end{tikzpicture}          
     \end{subfigure}
     \hfill
     \begin{tikzpicture}
        \begin{customlegend}[
            legend columns=2,
            legend style={
                align=center,
                column sep=4ex,
                font=\large,
            },
            legend entries={\longt, \slic}
        ]
        \addlegendimage{mark=\longtmark,mark size=3pt,solid,color=longtcolor}          
        \addlegendimage{mark=\modelmark,solid,color=sliccolor}
     
        \end{customlegend}
    \end{tikzpicture}            
    \caption{\textbf{\slic achieves stronger performance than \longt at any speed.} Average performance on all datasets as a function of inference and fine-tuning time per sample (ms) for \longt and \slic Base, Large, and XL models. \longt does not use MQA, but we report speed as though it had for a conservative baseline.}
    \label{fig:perf_vs_time}
\end{figure*}

In particular, \slic divides each feedforward layer and each attention layer into a {\em light branch} which is applied to all tokens and a {\em heavy branch} which is applied to a set of important tokens, selected specifically for that input and component. The light feedforward branch has lower hidden dimension than standard \longt while the heavy feedforward branch has higher hidden dimension. The light attention branch has fewer heads and applies only local attention, while the heavy attention branch performs full attention over another separately selected set of important tokens. Figure \ref{fig:slic} provides an overview of the \slic conditional mechanism.

Finally, \slic also includes two other modifications to the \longt architecture. \slic adds multi-query cross-attention~\citep{shazeer2019mq}, significantly speeding up inference. \slic also employs the UL2~\cite{tay2022ul2} pre-training objective, which we demonstrate allows for in-context learning over long inputs.


We show that \slic performs much faster fine-tuning and inference with similar or better model quality, improving over \longt on arXiv summarization~\citep{cohan2018arxiv} and TriviaQA question answering~\citep{joshi2017triviaqa} datasets and achieving SOTA on the SCROLLS benchmark~\citep{shaham2022scrolls}. Moreover, \slic achieves further gains in quality and speed for tasks with extremely long inputs (64k tokens), with less-than-linear scaling of ``focus'' tokens.

\section{Background}\label{sec:background}

\paragraph{Transformer FLOPs} \slic follows an extensive line of work in attempting to reduce the computational cost of Transformer models, particularly over long inputs. The computational burden of Transformer models has several distinct elements, and different approaches focus on reducing the cost of different components. For that reason, it is helpful to start by providing a breakdown of the computational cost of Transformer components. Table \ref{table:transformer_flops} shows the FLOPs\footnote{Each multiply-add is counted as a single FLOP.} for each component of a Transformer encoder layer~\citep{kaplan2020scaling}.

\begin{table}[h!]
\centering
\begin{tabular}{lc}
\toprule
\textbf{Encoder Layer Component} & \textbf{Flops} \\
\midrule
Vanilla self-attention computation& $2 n^2 d$ \\
Attention QKV and output projections & $4 n d^2$ \\
Feedforward layer& $8 n d^2$ \\
\longt local attention computation & $2 nwd$\\
\longt global attention computation & $\frac{n^2 }{8}d$\\
\bottomrule
\end{tabular}

\caption{Computational cost of encoder layer transformer components measured in FLOPs. $n$ is the input length, $d$ is the model dimensionality, and $w$ is the size of the local attention window.}
\label{table:transformer_flops}
\end{table}

\paragraph{Sparse attention} The first challenge of applying a Transformer to a long input is that the FLOPs of the self-attention mechanism scales quadratically in the input length, becoming intractable for long inputs. A large body of work focuses on reducing self-attention cost, restricting attention between a subset of inputs~\citep{child2019generating, ainslie2020etc, beltagy2020longformer, zaheer2020big, wang2020linformer, guo2022longt5} or to a subset of layers~\citep{zemlyanskiy2021readtwice}. In \longt~\citep{guo2022longt5}, the most closely related model to \slic, tokens attend within a local window as well as to a mean-pooled summary representation for each block of 16 tokens in the input. \longt attention leads to sharply reduced (though still non-negligible) FLOPs (Table \ref{table:transformer_flops}).

\paragraph{Conditional computation} After applying a sparse attention mechanism, the feedforward and attention projection layers account for the majority of the FLOPs. These costs scale with the length of the input, such that processing long inputs is still prohibitively expensive. A common approach to reduce the remaining cost is to employ some form of \textit{conditional computation}, avoiding applying all model parameters to the entire input. CALM~\cite{schuster2022confident} applies a varying number of decoder layers to each decoded token, outputting a token early if the model is confident in its prediction. Mixture-of-Experts models~\cite{shazeer2017moe, fedus2021switch, zoph2022stmoe} route inputs through a small proportion of expert sub-modules, bringing to bear only the parameters most relevant to the input. In the context of retrieval-augmented models, numerous works re-rank retrieved passages by their relevance to the query and process only the highest scoring passages~\citep{readerguidererank, r3rerank, kgfid} and vary the number of processed passages depending on model confidence~\citep{adaptiveretrieval, canext}. Concurrent work CoDA~\citep{lei2023conditional} employs a related conditional computation mechanism, designed for efficient adaptation rather than modeling long documents.

\paragraph{Device utilization} FLOPs do not tell the whole story, as modeling choices can influence the effective speed of operations achieved by accelerators. For long text inputs, autoregressive decoder inference is very slow due to memory bandwidth constraints from repeatedly loading the long sequence of keys and values~\citep{shazeer2019mq, dejong2022fido}. \citet{shazeer2019mq} introduces multi-query attention (MQA), sharing heads for keys and values to reduce memory bandwidth overhead. \citet{pope2022efficiently} studies how to shard large models, especially in the context of MQA, to obtain optimal device utilization and therefore speed.

\paragraph{Training objectives} T5 introduced the span corruption objective~\citep{raffel2020t5}, a modification of masked language modeling~\citep{devlinbert2019}. \longt made use of the PEGASUS~\cite{zhang2020pegasus} sentence reconstruction objective for improved summarization performance. \citet{tay2022ul2} proposes UL2, a mixture of span corruption, prefix, and causal language modeling, and shows that it leads to strong performance on both short-output and generative tasks.

\section{\slic}\label{sec:approach}

\subsection{Conditional computation}

As discussed in the previous section, a large proportion of Transformer FLOPs arise from feedforward and projection layers that scale with the length of the input sequence. Therefore, \longt training and inference on long documents remains expensive. 

\slic further reduces the cost of processing long documents through \textit{conditional computation}, following the intuition that some tokens are more important and therefore benefit more than others from heavy computation. First, some types of tokens may inherently require less computation, such as filler words and punctuation. Second, especially in long documents, large parts of the input may not be relevant to the current question, task, or processing stage.

The \slic conditional computation mechanism consists of three components: routing modules, conditional feedforward layers, and conditional attention layers. All tokens are processed by standard, lightweight attention and feedforward layers. Routing modules additionally select important tokens from an input at each attention or feedforward layer, and a heavy conditional layer applies additional computation to routed tokens. This section describes each component in detail. Figure~\ref{fig:slic} provides an overview of the \slic conditional computation mechanism, and Table \ref{table:total_transformer_flops} compares \slic and \longt FLOPs.

\begin{table}[h!]
\centering
\begin{tabular}{ll}
\toprule
\textbf{Model} & \textbf{Encoder Layer Flops} \\
\midrule
\textsc{T5}& $12 n d^2 + 2n^2 d$ \\
\longt & $12n d^2 + \frac{n^2}{8} d$ \\
\slic& $7 \frac{1}{4} n d^2 + \frac{n^2}{84} d$ \\
\bottomrule
\end{tabular}
\caption{\textbf{\slic uses significantly fewer FLOPs than \longt.} Comparison of approximate encoder layer total FLOPs between \textsc{T5}, \longt, and \slic. \slic FLOPs rounded to readable fractions.}
\label{table:total_transformer_flops}
\end{table}

\paragraph{Routing}

In order to separately select important tokens for each component in each layer, we need a \textit{learnable} and \textit{tractable} routing function. We follow the simple three-step mechanism from \citet{lei2023conditional}: (1) multiply inputs with a learned embedding to obtain routing scores, (2) normalize, and (3) select the top-$k$ highest scoring inputs.

Let $X_i$ be the representation of token $i$, and $u$ a $d$-dimensional learnable embedding. Then the routing score of token $i$ is 
\begin{equation*}
 s_i = X_i \cdot u
\end{equation*}
We select the top-$k$ highest scoring inputs. In order to provide a learning signal to the scoring embedding, we make sure the contribution of the routed tokens to the layer update is \textit{scaled} according to the routing score, as will be seen later. To provide a better distributed signal to all tokens, we also globally normalize the routing scores to sum up to the number of desired routed tokens using a generalized softmax, resulting in normalized scores $\tilde{s}_i$.
Each \slic layer has three independent routers, one each for the feedforward layer, attention queries, and attention key-values.

\paragraph{Conditional Feedforward}
Intuitively, some token representations may benefit from more processing than others. The \slic conditional feedforward layer applies an additional high-capacity feedforward layer to selected tokens. In particular, let $X_i$ be the model state of the $i$th token and $\tilde{s}_i$ denote the normalized routing score (set to 0 for non-routed tokens). Then the feedforward update for \slic is given by
\begin{equation*}
    X_i = X_i + \textrm{FFd}_{\textrm{Light}}(X_i) + \tilde{s}_i \cdot \textrm{FFd}_{\textrm{Heavy}}(X_i)
\end{equation*}
The light and heavy feedforward branches differ only in their hidden dimension, with the light branch having smaller hidden dimension than the standard T5 feedforward layer and the heavy branch larger. Let $n$ denote the number of input tokens, $m$ the number of selected tokens, and $r_L$ and $r_H$ the ratios of light and heavy hidden dimension to standard T5 hidden dimension. Then the FLOPs of the \slic layer are given by
\begin{equation*}
    \text{FLOPs}_{\text{FFd}} = \underbrace{8 n r_L d^2}_{\text{Light branch}} + \underbrace{8 m r_H d^2}_{\text{Heavy branch}}
\end{equation*}
We set the light and heavy ratios as $r_L = \frac12$ and $r_H = 4$, half and quadruple the standard T5 hidden dimension respectively. For our main experiments, a fraction $\frac{1}{16}$ of tokens are routed to the heavy branch. As a result the approximate FLOPs from the \slic feedforward layer equals 
\begin{equation*}
    \text{FLOPs}_{\text{FFd}} = \underbrace{4 n d^2}_{\text{Light branch}} + \underbrace{2 n d^2}_{\text{Heavy branch}}
\end{equation*}
consuming 75\% of the FLOPs of a standard T5 feedforward layer.
\begin{table*}[t!]
\centering
\footnotesize
\begin{tabular}{l|c|cc|ccccccccc}
\toprule
\textbf{Model} & \textbf{Avg} & \multicolumn{2}{|c|}{\textbf{Speed}}& \textbf{TQA} & \textbf{NQA} &  \textbf{QAS} & \textbf{QuAL}  &\textbf{CNLI} & \textbf{arXiv} &  \textbf{SumS} & \textbf{QMS}  & \textbf{GovR}    \\
\midrule
 &  & \textbf{inf} & \textbf{fn} & \textbf{F1} & \textbf{F1} &  \textbf{F1} & \textbf{EM}  &\textbf{EM} & \textbf{R\textsubscript{gm}} &  \textbf{R\textsubscript{gm}} & \textbf{R\textsubscript{gm}}  & \textbf{R\textsubscript{gm}}    \\
\midrule
 \longt-B & 43.1 & 0.6 / 7.4&3.7   & 82.2 & 23.0 & 46.6 & 37.9 & 85.6 &  35.4 & 19.2 & 20.4 &  37.7  \\    
\slic-B & 42.4 & 11.2&6.5 & 82.4    & 23.3 & 42.1 & 36.5   & 86.5 & 35.3 & 18.7 & 18.4 & 37.9\\ 
\midrule
\longt-L  & 45.3 & 0.3 / 3.0&1.3  & 84.2 & 27.2 & 52.3 & 40.6 & 87.3 &  35.7 & 19.1 & 21.4 &  39.5  \\    
 \slic-L  & 45.3 & 5.0&2.0  & 84.5 & 27.7 & 49.8 &39.9 & {\bf 88.7} &  35.9 & {\bf 20.5} & 21.0 &  39.7  \\ 
\midrule
\longt-XL  & 46.6 & 0.2 / 1.2 &0.4  & 85.3 & 29.3 & 53.1 & 46.0 & 88.2 &  35.9 & 19.4 & 21.3 &  \textbf{40.5}  \\    
 \slic-XL & \textbf{47.4} & 2.3 & 0.5   & \textbf{86.1} & \textbf{31.1} & \textbf{53.9} & \textbf{48.1} & 88.4 & \textbf{36.1} & 20.0 & \textbf{22.5} &  \textbf{40.5}  \\  
    \bottomrule
\end{tabular}
\caption{Performance comparison of \slic and \longt Base, Large and XL models on question-answering datasets TriviaQA (TQA), NarrativeQA (NQA), QASPER (QAS), and QuALITY (QuAL), NLI dataset ContractNLI (CNLI), and summarization datasets arXiv, SummScreenFD (SumS), QMSum (QMS), and GovReport (GovR). SCROLLS results are on leaderboard test set where \slic-XL achieves SOTA. Average speed is reported in samples per second for inference (inf) and fine-tuning (fn). \longt does not use MQA but inference speed is reported without/with MQA for conservative baseline. R\textsubscript{gm} stands for the geometric mean of ROUGE-1,2,L. Similar to SCROLLS, we take a simple average across all datasets even though the datasets use different performance metrics.}
\label{table:headline_results}
\end{table*}
\paragraph{Conditional Attention}
\begin{figure}[t!]
    \centering
    \includegraphics[width=0.95\columnwidth]{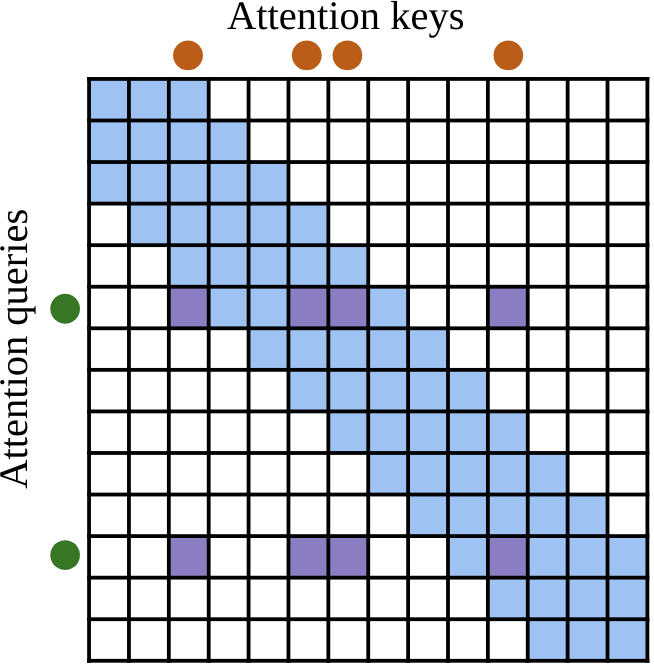}
    \caption{An overview of the \slic attention pattern. The light branch performs local attention for each token. In the higher capacity heavy branch $q$ selected query tokens (2 in the figure) attend to $v$ separately selected key and value tokens (4 in the figure).}
    \label{fig:slic_attention}
\end{figure}
\slic conditional attention operates on the intuition that most tokens have simple, local interactions, but some tokens benefit from heavier processing and long-range interactions. The \slic conditional attention layer applies an additional high-capacity attention layer that attends from selected query tokens to selected key-value tokens. Let $\tilde{s}_i^{q}$ denote the normalized routing query score for token $i$, and $\tilde{s}^{kv}$ the key-value scores for all tokens (set to $0$ if not routed). Then the attention update for \slic is given by
\begin{equation*}
    X_i = X_i + \textrm{A}_{\textrm{Light}}(X_i, X) + \tilde{s}_i^{q} \cdot \textrm{A}_{\textrm{Heavy}}(X_i, \tilde{s}^{kv}X)
\end{equation*}
The light and heavy branches differ in the number of heads and tokens attended to: the light branch has fewer heads and attends to a local context window, while the heavy branch has more heads and attends to all routed key-value tokens. Separately selecting query and key-value tokens also allows the model to differentiate between tokens that \textit{require} additional information and those that \textit{possess} such information. Figure \ref{fig:slic_attention} shows the \slic attention pattern. Let $q, v$ be the number of selected query and key-value tokens, $w$ the size of the local attention window and $r_L, r_H$ the proportion of light and heavy heads relative to standard T5. Then the FLOPs of the \slic  attention layer are given by
\begin{align*}
    \text{FLOPs}_{\text{Att}} &= \underbrace{4 n \cdot r_L d^2}_{\text{Local projection}} + \underbrace{2 n  w \cdot r_L d}_{\text{Local attention}} \\ &+ \underbrace{2 q \cdot r_H d^2 + 2 v \cdot r_H d^2}_{\text{Global projection}} + \underbrace{2 q v \cdot r_{H} d}_{\text{Global attention}}
\end{align*}
We set the light and heavy head ratios as $r_L = \frac14$ and $r_H = \frac34$, keeping the total number of heads across the light and heavy branches equal to standard T5 heads. For our main experiments a fraction $\frac{1}{16}$ query tokens and $\frac{1}{8}$ key-value tokens are routed to the heavy branch, so $q = \frac{n}{16}$ and $v = \frac{n}{8}$. Ignoring local attention computation, we approximate attention FLOPS by\footnote{Global projection and attention FLOPs rounded to readable fractions, exact values are $\frac{9}{32}$ and $\frac{3}{256}$. Complexity assumes constant fraction of routed tokens; we show we can do better in practice for extremely long inputs.}

\begin{equation*}
    \text{FLOPs}_{\text{Att}} \approx \underbrace{n d^2}_{\text{Local proj.}} + \underbrace{\frac{1}{4}n d^2}_{\text{Global proj.}}  + \underbrace{\frac{1}{84}n^2 d}_{\text{Global att.}}
\end{equation*}
with less than half projection FLOPs and order-of-magnitude smaller quadratic length scaling compared to \longt. Table \ref{table:total_transformer_flops} shows total FLOPs for the \slic layer. In general, we set $q = m$ and $v = 2m$, and use $m$ to summarize the number of routed tokens going forward.

\subsection{Multi-query Attention}

Conditional computation effectively reduces the computational cost of the encoder. However, for encoder-decoder models with long inputs the majority of inference time is spent in the decoder due to memory bandwidth constraints~\citep{shazeer2019mq, dejong2022fido}. Most of the overhead is caused by repeatedly reading all the input token keys and values from memory for every output token that is autoregressively decoded during cross attention. Multi-query attention~\citep{shazeer2019mq} (MQA) allows all query heads to share a single key and value head, alleviating this bottleneck. Accordingly, we apply MQA in cross-attention layers for much faster inference. Note however that MQA does not improve training speed since target tokens are processed in parallel during training, avoiding this memory bandwidth bottleneck.

\subsection{UL2}

The UL2 pre-training objective~\citep{tay2022ul2} combines different denoising objectives, extending the span corruption pre-training used in T5 to a variety of noise rates / average span lengths and adding a prefix language modeling objective more similar to typical decoder-only model pre-training. UL2 has been shown to lead to improved in-context learning. We train \slic on UL2 instead of PEGASUS~\cite{zhang2020pegasus}, endowing \slic with in-context learning capabilities.

\section{Experiments}\label{sec:experiments}

In order to evaluate \slic, we perform the following experiments: (1) our main results compare \slic and \longt on a collection of long input datasets using input length of 16k tokens; (2) we evaluate \slic on extremely long inputs up to 64k tokens and compare scaling against \longt; (3) demonstrate \slic's few-shot capability, investigating how performance changes as input length and number of shots increase, (4) perform a series of ablations to understand the effect of individual \slic components, and (5) investigate empirical routing patterns. The remainder of the section outlines our experimental setup, and then describes each of the experiments above.

\subsection{Experimental setup}

\paragraph{Configurations}

\slic is based on the T5.1.1 architecture ~\citep{raffel2020t5}, implemented with JAX~\citep{jax}, Flax~\citep{flax}, and Flaxformer\footnote{https://github.com/google/flaxformer}. Following \longt, we experiment with Base, Large, and XL model sizes. \slic models use the same embedding dimension, number of layers, and total attention heads as corresponding \longt models of the same size, with more overall parameters (but less compute) due to the conditional branch. See Appendix \ref{sec:appendix-parameters} for additional details on model configuration.

\paragraph{Pre-training}
We pre-train \slic for 1M steps on the C4 dataset~\citep{raffel2020t5} using a variant of the UL2 objective~\citep{tay2022ul2} with batch size 256, input length 4096, and output length 910. In particular, our mixture contains four objectives in equal proportion: prefix-LM with noise rate 0.5, and span corruption~\citep{raffel2020t5} with noise rate 0.15 and average span lengths 3, 8, and 64. We use the Adafactor optimizer~\citep{adafactor} with the T5.1.1 inverse square root learning rate schedule and no dropout. \slic is trained with the T5X~\citep{t5x} framework. For pre-training, we route $m = 512$ tokens, $\frac{1}{8}$th of the input length.

\paragraph{Fine-tuning}
For fine-tuning we use a constant learning rate of 0.001, batch size 128, and dropout rate 0.1 for all tasks. Main results use input length of 16384 for all datasets other than ContractNLI, which uses 8192. Question answering datasets use output length 128 and summarization datasets use output length 512, except for GovRep which uses output length 1024. We route $m = 1024$ tokens, $\frac{1}{16}$th of the input length. We train until convergence and select the checkpoint with the highest dev performance. We use greedy decoding for inference.

\paragraph{Data}

We evaluate \slic on TriviaQA~\citep{joshi2017triviaqa}, arXiv~\citep{cohan2018arxiv}, and the SCROLLS benchmark~\citep{shaham2022scrolls}. SCROLLS contains question-answering datasets:  NarrativeQA~\citep{kocisky2018narrativeqa}, QASPER~\citep{dasigi2021qasper}, and QuALITY~\citep{pang2021quality}, an NLI dataset: ContractNLI~\citep{koreeda-manning-2021-contractnli}, and summarization datasets: SummScreenFD~\citep{chen-etal-2022-summscreen}, QMSum~\citep{zhong2021qmsum}, and GovReport~\citep{huang2021govreport}. Table \ref{table:data_stats} provides an overview of the size and input length for each dataset.

\begin{table}[ht!]
\small
\centering
\vspace{0.35cm}
\begin{tabular}{l|cccc}
    \textbf{Dataset} & \textbf{Type} & \textbf{Samples} & \textbf{Median} & \textbf{90\%}\\
    \toprule
    TriviaQA & QA &  157,053 & 8,858 & 28,956 \\   
    arXiv & Sum & 215,913 & 8,519 & 20,170 \\   
    NarrativeQA & QA & 71,187 & 57,829 & 176,862 \\    
    QASPER & QA & 5,692 & 5,472 & 8,657 \\    
    QuALITY & QA & 6,737 & 7,171 & 8,276 \\   
    ContractNLI & NLI & 10,319 & 2,148 & 4,485 \\    
    SummScreen & Sum & 4,348 & 9,046 & 15,172 \\    
    QMSum & Sum & 1,810 & 14,197 & 27,761 \\   
    GovRep & Sum  & 19,402 & 8,841 & 18,835 \\    
    \bottomrule
\end{tabular}
\caption{Median and 90th percentile input length by dataset measured in SentencePiece tokens.}
\label{table:data_stats}
\end{table}

\paragraph{Timing}

We report time per sample per TPUv4 chip, as measured by xprof \citep{xprof}. For inference we use a single TPUv4 with batch size 16 or the largest that fits in memory. For fine-tuning we profile with 8 TPUv4 chips, sharded separately for each model to maximize throughput.

\subsection{Main results}

Figure \ref{fig:perf_vs_time} compares the quality-speed trade-off for \longt\footnote{Note that \longt does not use MQA, but for profiling we add MQA to \longt for a conservative baseline.} and \slic, showing that \slic is better at any speed. For 16k input length, \slic matches or exceeds \longt quality for Large and XL with 35-75\% training speedup and 50-100\% inference speedup on top of the order-of-magnitude inference speedup from MQA. Encoder speedups are even greater (Appendix \ref{sec:appendix-results}). \slic-XL also achieves SOTA performance on the SCROLLS benchmark. Table \ref{table:headline_results} contains all main results.

\subsection{Scaling to extremely long inputs}
\begin{figure}
     \centering
        \begin{tikzpicture}[scale=1.0]
            \begin{axis}[
            scale only axis,
            width=0.8\columnwidth,
            ylabel={F1},
            xlabel={Time per sample (ms)},
            mark=x,
            ymajorgrids=true,
            xmajorgrids=true,
            xminorticks=true,
            grid style=dashed,
            legend columns=1,
            legend cell align=left,
            legend pos={south east},
        ]
            \addplot[color=longtcolor,mark=\longtmark,mark size=2pt, line width=2] table {
                100 23.04
                222 25.57
                613 28.43
            };        
            \addplot[color=sliccolor,mark=\modelmark,mark size=2pt, line width=2] table {
                69 23.62
                103 26.69
                262 29.87
                473 32.20
                			
            };
            \node at (axis cs:100,23.04) [anchor= west, color=darkgrey] {\small{8k}};   
            \node at (axis cs:200,25.57) [anchor= north west, color=darkgrey] {\small{16k}};            
            \node at (axis cs:613,28.35) [anchor= north, color=darkgrey] {\small{32k}};            
            
            \node at (axis cs:69,23.62) [anchor= west, color=darkgrey] {\small{8k}};   
            \node at (axis cs:103,26.69) [anchor= west, color=darkgrey] {\small{16k}};            
            \node at (axis cs:248,29.74) [anchor= north west, color=darkgrey] {\small{32k}};            
            \node at (axis cs:435,32.0) [anchor= north west, color=darkgrey] {\small{64k}};            
            
             \legend{\longt, \slic}        
            \end{axis}
        \end{tikzpicture}
    \caption{\textbf{\slic effectively scales to extremely long inputs, achieving stronger performance and faster speed than \longt.} F1 on NarrativeQA as a function of inference time per sample for \longt and \slic Large models using varying input lengths.}
    \label{fig:perf_vs_time_len}
\end{figure}
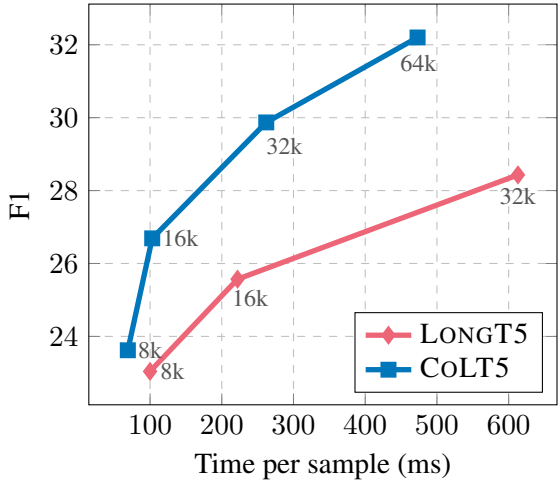
We hypothesize that the advantage of \slic over \longt strengthens with input length, as the fraction of important tokens decreases and \slic can route a greater proportion of important tokens to the heavy branch. Figure \ref{fig:perf_vs_time_len} compares the quality-speed trade-off for \longt and \slic on NarrativeQA, sweeping over input length rather than model size. The number of routed tokens is $\frac{1}{16}$th of the input length, except that we do not increase routed tokens going from 32k to 64k, so at 64k we route only $\frac{1}{32}$nd of the input length. \slic achieves both stronger performance and faster inference speed at all input lengths and is able to effectively make use of extremely long inputs. We note that \slic achieves large quality gains by going from 32k to 64k tokens even while keeping the number of routed tokens constant, providing more evidence for our hypothesis. 

\subsection{In-context learning}

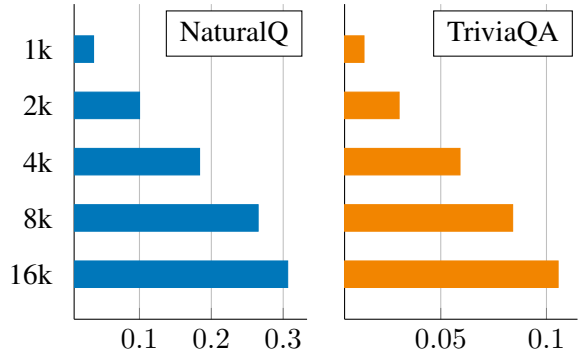
\begin{figure}[h!]
\begin{subfigure}[t!]{0.24\textwidth}
\hspace{-10pt}
\begin{tikzpicture}[baseline={(current bounding box.south)},scale=1.0]
\begin{axis}[
    xbar,
    enlarge y limits=0.2,
    width=1.21\columnwidth,
    height=1.5\columnwidth,
    major y tick style = transparent,    
    xmajorgrids = true,
    symbolic y coords={1k, 2k, 4k, 8k, 16k},    
    ytick = data,
    y dir=reverse,
    axis y line*=none,
    axis x line*=bottom,
]
    \addlegendimage{empty legend}\addlegendentry{NaturalQ}
    \addplot[style={fewnqcolor,fill=fewnqcolor,mark=none}]
        coordinates {(0.0361,1k) (0.1001,2k) (0.1836,4k) (0.2651,8k) (0.3062,16k)};
\end{axis}
\end{tikzpicture}            
\end{subfigure} \hspace{-4pt}
\begin{subfigure}[t!]{0.24\textwidth}
\begin{tikzpicture}[baseline={(current bounding box.south)},scale=1.0]
\begin{axis}[
    xbar,
    enlarge y limits=0.2,
    width=1.21\columnwidth,
    height=1.5\columnwidth,
    major y tick style = transparent,    
    xmajorgrids = true,
    symbolic y coords={1k, 2k, 4k, 8k, 16k},  
    yticklabels={},    
    ytick = data,
    y dir=reverse,
    axis y line*=none,
    axis x line*=bottom,
    /pgf/number format/fixed,
]
    \addlegendimage{empty legend}\addlegendentry{TriviaQA}
    \addplot[style={fewtqacolor,fill=fewtqacolor,mark=none}]
        coordinates {(0.0137,1k) (0.0303,2k) (0.0591,4k) (0.0840,8k) (0.1055,16k)};

\end{axis}
\end{tikzpicture}            
\end{subfigure}
\caption{\textbf{\slic can use its long-input capability to benefit from more shots for in-context learning.} Few-shot exact match for \slic-Large on Natural Questions and TriviaQA dev sets as a function of input tokens, fitting as many examples as possible. Each example contains question, context, and answer. Inputs length used are 1024, 2048, 4096, 8192, 16384.}
\label{fig:few_shot}
\end{figure}

Models trained on the UL2 objective have shown strong few-shot in-context learning (ICL) capabilities\footnote{We initially evaluated ICL for models pre-trained with PEGASUS but found performance to be nearly 0.} even at smaller sizes~\citep{tay2022ul2}. \slic enables tractable inference with long inputs. Here, we leverage this for scaling the number of examples used for in-context learning. 

We test the above hypothesis by evaluating few-shot learning performance on Natural Questions~\citep{kiwatkowski2019nq} and TriviaQA as a function of input length, using as many examples as fit in the context. We consider the open book setting, such that each example consists of question, context document, and answer. Table \ref{table:few_shot_n_shots} shows the number of examples by input length. We evaluate on the full dev set, randomly sampling examples from the training set for each dev sample until no further examples fit in the input length. We found that \slic can perform in-context learning only up to the input length it was trained on, so for these experiments we continued pre-training a \slic-Large model on input length 16384 for another 100k steps. For the same reason we route $m = 512$ tokens as in pre-training.

Figure \ref{fig:few_shot} displays \slic few-shot performance as a function of input length, showing that \slic is able to apply its long-input capabilities to extract information from increasing numbers of examples.

\begin{table}[h]
\small
\centering
\vspace{0.35cm}
\begin{tabular}{l|ccccc}
    \textbf{Dataset} & \textbf{1024} & \textbf{2048} & \textbf{4096} &  \textbf{8192} & \textbf{16384} \\
    \toprule
    NQ & 0.1 & 0.7 & 1.7 & 3.4 & 5.6 \\
    TriviaQA & 1.6 & 2.3 & 3.8 & 7.0 & 9.8 \\
    \bottomrule
\end{tabular}
\caption{Average number of Natural Questions and TriviaQA few-shot examples that fit in input length.}
\label{table:few_shot_n_shots}
\end{table}

\subsection{Ablations}
\begin{table*}[t!]
\small
\centering
\begin{tabular}{ll|cc|ccccccccc}
\toprule
\multirow{2}{*}{\textbf{Ablation}} & \multirow{2}{*}{\textbf{Model}} & \textbf{Avg} & \textbf{Inf} & \textbf{TQA} & \textbf{NQA} &  \textbf{QAS} & \textbf{QuAL}  &\textbf{CNLI} & \textbf{arX} &  \textbf{SumS} & \textbf{QMS}  & \textbf{GovR}  \\
\cmidrule{3-13}
 & &  & \textbf{S/s} & \textbf{F1} & \textbf{F1} &  \textbf{F1} & \textbf{EM}  &\textbf{EM} & \textbf{R\textsubscript{gm}} &  \textbf{R\textsubscript{gm}} & \textbf{R\textsubscript{gm}}  & \textbf{R\textsubscript{gm}}    \\
\midrule
Baseline & \slic-B    & 42.5 & 11.2 & 82.4    & 23.1 & 38.3 & 36.6   & 87.8 & 35.3 & 19.3 & 20.5 & 39.4\\
\toprule
\multirow{ 2}{*}{Routing} & Static    & 40.5 & 11.6 & 79.7    & 19.2 & 34.2 & 34.5   & 86.4 & 34.9 & 18.1 & 18.9 & 38.8\\
 & Share QKV    & 42.0 & 11.8 & 82.1    & 21.9 & 37.5 & 36.2   & 87.0 & 35.2 & 18.2 & 20.4 & 39.7 \\
\midrule
\multirow{ 2}{*}{Attention} & v=all    & 42.5 & 9.4 & 82.4 & 22.3    & 38.6 & 37.2 & 87.8   & 35.3 & 19.1 & 20.3 & 39.8  \\
 & v=q    & 42.3 & 11.5 & 82.5    & 22.5 & 37.3 & 37.0   & 85.9 & 35.2 & 19.0 & 20.5 & 39.7 \\
 \midrule
\multirow{ 2}{2pt}{Routed Tokens} & m=512    & 41.6 & {\bf 12.2} & 81.9    & 22.1 & 37.3 & 35.4   & 84.6 & 35.2 & 18.9 & 19.5 & 39.6 \\
 & m=1536    & {\bf 42.9} & 10.4 & 82.6    & {\bf 23.5} & 39.8 & {\bf 37.5}   & 87.5 & 35.4 & 19.4 & 20.8 & 40.0\\
\midrule
Encoder & \longt-B & 42.1 & 7.4 & 82.0    & 21.4 & 38.4 & 35.8   & {\bf 88.0} & {\bf 35.5} & 18.7 & 20.4 & 38.5\\ 
\midrule
Decoder & Multi-head & {\bf 42.9} & 0.7 & {\bf 82.7}    & 22.9 & 40.2 & 35.8   & 87.7 & {\bf 35.5} & {\bf 19.7} & {\bf 21.2} & {\bf 40.3}\\
\midrule
Objective & PEGASUS     & 42.8 & 11.2 & 82.6    & 22.6 & {\bf 40.5} & 37.3   & 87.3 & 35.3 & 19.6 & 20.8 & 39.6\\

    \bottomrule
\end{tabular}
\caption{\slic ablations evaluated on validation sets. Each experiment modifies a component of the \slic recipe for \slic-Base. Static routing divides the input into equal-length blocks and selects the first token in each block to be routed. Shared QKV routing shares routing decisions for queries and keys/values. In v=all the routed queries attend to the entire input, while v=q selects the same number of key and value tokens as query tokens. m=512 and m=1536 use different numbers of routed tokens. \longt-B uses a \longt encoder while retaining other parts of the \slic training recipe such as MQA and the UL2 objective. Multi-head refers to using multi-head cross-attention. The final ablation replaces the UL2 objective with PEGASUS as in \longt.}
\label{table:ablations}
\end{table*}
\vspace{-0.1cm}

This section studies the effect of different choices in the \slic recipe. Table \ref{table:ablations} contains results of a series of experiments that change a single component for \slic Base. 

\paragraph{Routing} First, we note that static routing -{}- evenly distributing routed tokens over the input -{}- leads to massive drop in performance. The importance of routing provides evidence that the model learns to devote capacity to important tokens and the advantage of \slic is not merely a result of additional parameters. Sharing routing decisions for query and KV tokens should be compared with v=q, and leads to a modest reduction in quality and increase in speed.

The optimal number of routed tokens represents a trade-off between improved performance and computational cost of applying heavier layers. Table \ref{table:ablations} shows strong gains going from 512 to 1024 (baseline) routed tokens and diminishing returns for further increases.

\paragraph{Attention} 

\slic relies on routing to identify not only tokens that can benefit from important information elsewhere in the input, but also which tokens contain such important information. We study whether \slic is successful in this task by comparing performance with two different attention settings -{}- v=all, in which routed tokens attend to the entire input, and v=q, which uses equal number of routed keys and values as queries, rather than twice as many. \slic appears to occupy a sweet spot, as using fewer routed key-values modestly decreases performance at similar speed but attending to all inputs barely helps at sharply increased cost.

\paragraph{Other} 

We compare \slic to \longt with multi-query cross-attention, confirming that \longt indeed does not achieve an unexpected quality gain from MQA, and our conservative assumptions in Figures \ref{fig:perf_vs_time}, \ref{fig:perf_vs_time_len} are valid. Next, we evaluate multi-head cross-attention for \slic, finding that it leads to modestly improved \slic performance. However, as MHA exhibits order-of-magnitude slower inference, MQA is clearly favored. Finally, PEGASUS appears to fine-tune slightly better than UL2, though the difference is small and UL2 enables few-shot learning.

\subsection{Routing analysis}

\begin{figure}
\begin{tikzpicture}
\begin{axis}[
    ybar,
    bar width=.4cm,
    width=0.95\columnwidth,
    height=0.7\columnwidth,
    ylabel={Routed proportion},
    symbolic x coords={MLP, Query, KV},
    xtick=data,
    enlarge x limits=0.25,
    legend style={at={(0.25,0.95)},
      anchor=north,legend columns=1}
    ]    
    ]
    \addplot[fill=othercolor] coordinates {(MLP,0.212) (Query,0.211) (KV,0.343)};
    \addplot[fill=answercolor] coordinates {(MLP,0.236) (Query,0.262) (KV,0.432)};
    \addplot[fill=questioncolor] coordinates {(MLP,0.396) (Query,0.365) (KV,0.608)};
    \legend{Other, Answer, Question}
\end{axis}
\end{tikzpicture}
    \caption{Proportion of tokens routed for answer (string match), question, and other tokens by routing component for \slic Large model, averaged over examples in TriviaQA dev set and all layers of model.}
    \label{fig:routing_prop}
\end{figure}
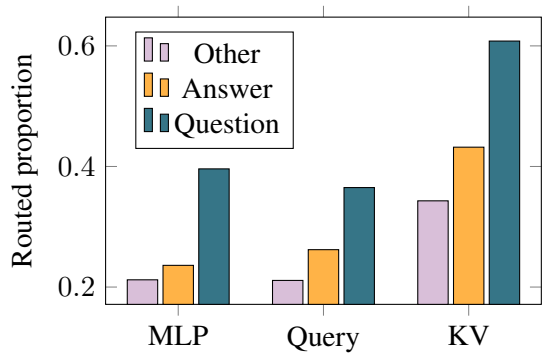

It is interesting to ask whether \slic routed tokens line up with what we consider intuitively important tokens in each document. We investigate this question by studying routing patterns of a Large \slic model fine-tuned on TriviaQA. We divide tokens into three categories: (1) question tokens, (2) answer tokens, and (3) other tokens. Figure \ref{fig:routing_prop} shows the average fraction of each type of token that is routed through the heavy path for MLP and attention layers on TriviaQA. We note that question and answer tokens are significantly more likely to be routed than other tokens, for feedforward as well as attention queries and keys/values. Appendix \ref{sec:appendix-routing} presents more detailed routing analysis; e.g., semantically important tokens are much more likely to be selected in later layers.

\section{Conclusion}\label{sec:conclusions}

We propose \slic, a new model for long-range inputs that employs conditional computation for higher quality and faster speed. \slic has light feedforward and attention layers that apply to the entire input, as well as heavy branches that are applied only to a subset of important tokens selected by a learned router. We show that \slic achieves stronger performance at any speed compared to \longt on a variety of long-input datasets, and can effectively and efficiently make use of extremely long inputs up to 64k tokens.

\section*{Limitations}

\slic applies conditional computation only in the encoder. Applying conditional computation in the decoder is more complicated; the routing method in \slic is not causal, so it isn't applicable when generating token by token. Since decoder-only models and applications with long outputs have become more popular recently, this is a strong limitation of the current approach. Although the routing method in \slic could potentially be applied to the \emph{input} context in a decoder-only model, we didn't investigate this setup.

\slic is specialized towards long sequences and has to be trained from scratch. For large-scale training and deployment, it is desirable to either train a single model that can handle both short and long sequences, or develop a long-input architecture that can be adapted from an existing large model. 

\section*{Acknowledgements}

We would like to thank Srinadh Bhojanapalli, Luke Vilnis, Zachary Fisher, Jianmo Ni, Tal Schuster, Vaclav Cvicek, Sudeep Gandhe, Bhargav Kanagal, Kenton Lee, Ming-Wei Chang, Afroz Mohiuddin, Raphael Hoffmann, and others at Google Research for helpful advice and discussion.

\bibliography{anthology,custom}

\begin{thebibliography}{43}
\expandafter\ifx\csname natexlab\endcsname\relax\def\natexlab#1{#1}\fi

\bibitem[{Ainslie et~al.(2020)Ainslie, Onta\~{n}\'{o}n, Alberti, Cvicek,
  Fisher, Pham, Ravula, Sanghai, Wang, and Yang}]{ainslie2020etc}
Joshua Ainslie, Santiago Onta\~{n}\'{o}n, Chris Alberti, Vaclav Cvicek, Zachary
  Fisher, Philip Pham, Anirudh Ravula, Sumit Sanghai, Qifan Wang, and Li~Yang.
  2020.
\newblock \href {https://arxiv.org/abs/2004.08483} {{ETC}: Encoding long and
  structured inputs in transformers}.
\newblock \emph{arXiv preprint arXiv:2004.08483}.

\bibitem[{Beltagy et~al.(2020)Beltagy, Peters, and
  Cohan}]{beltagy2020longformer}
Iz~Beltagy, Matthew~E Peters, and Arman Cohan. 2020.
\newblock Longformer: The long-document transformer.
\newblock \emph{arXiv preprint arXiv:2004.05150}.

\bibitem[{Bradbury et~al.(2018)Bradbury, Frostig, Hawkins, Johnson, Leary,
  Maclaurin, Necula, Paszke, Vander{P}las, Wanderman-{M}ilne, and Zhang}]{jax}
James Bradbury, Roy Frostig, Peter Hawkins, Matthew~James Johnson, Chris Leary,
  Dougal Maclaurin, George Necula, Adam Paszke, Jake Vander{P}las, Skye
  Wanderman-{M}ilne, and Qiao Zhang. 2018.
\newblock \href {http://github.com/google/jax} {{JAX}: composable
  transformations of {P}ython+{N}um{P}y programs}.

\bibitem[{Chen et~al.(2022)Chen, Chu, Wiseman, and
  Gimpel}]{chen-etal-2022-summscreen}
Mingda Chen, Zewei Chu, Sam Wiseman, and Kevin Gimpel. 2022.
\newblock \href {https://doi.org/10.18653/v1/2022.acl-long.589}
  {{S}umm{S}creen: A dataset for abstractive screenplay summarization}.
\newblock In \emph{Proceedings of the 60th Annual Meeting of the Association
  for Computational Linguistics (Volume 1: Long Papers)}, pages 8602--8615,
  Dublin, Ireland. Association for Computational Linguistics.

\bibitem[{Child et~al.(2019)Child, Gray, Radford, and
  Sutskever}]{child2019generating}
Rewon Child, Scott Gray, Alec Radford, and Ilya Sutskever. 2019.
\newblock Generating long sequences with sparse transformers.
\newblock \emph{arXiv preprint arXiv:1904.10509}.

\bibitem[{Cohan et~al.(2018)Cohan, Dernoncourt, Kim, Bui, Kim, Chang, and
  Goharian}]{cohan2018arxiv}
Arman Cohan, Franck Dernoncourt, Doo~Soon Kim, Trung Bui, Seokhwan Kim, Walter
  Chang, and Nazli Goharian. 2018.
\newblock \href {https://doi.org/10.18653/v1/N18-2097} {A discourse-aware
  attention model for abstractive summarization of long documents}.
\newblock In \emph{Proceedings of the 2018 Conference of the North {A}merican
  Chapter of the Association for Computational Linguistics: Human Language
  Technologies, Volume 2 (Short Papers)}, pages 615--621, New Orleans,
  Louisiana. Association for Computational Linguistics.

\bibitem[{Dasigi et~al.(2021)Dasigi, Lo, Beltagy, Cohan, Smith, and
  Gardner}]{dasigi2021qasper}
Pradeep Dasigi, Kyle Lo, Iz~Beltagy, Arman Cohan, Noah~A. Smith, and Matt
  Gardner. 2021.
\newblock \href {https://doi.org/10.18653/v1/2021.naacl-main.365} {A dataset of
  information-seeking questions and answers anchored in research papers}.
\newblock In \emph{Proceedings of the 2021 Conference of the North American
  Chapter of the Association for Computational Linguistics: Human Language
  Technologies}, pages 4599--4610, Online. Association for Computational
  Linguistics.

\bibitem[{de~Jong et~al.(2022)de~Jong, Zemlyanskiy, Ainslie, FitzGerald,
  Sanghai, Sha, and Cohen}]{dejong2022fido}
Michiel de~Jong, Yury Zemlyanskiy, Joshua Ainslie, Nicholas FitzGerald, Sumit
  Sanghai, Fei Sha, and William Cohen. 2022.
\newblock \href {https://arxiv.org/abs/2212.08153} {Fi{DO}: Fusion-in-decoder
  optimized for stronger performance and faster inference}.
\newblock \emph{arXiv preprint arXiv:2212.08153}.

\bibitem[{Devlin et~al.(2019)Devlin, Chang, Lee, and
  Toutanova}]{devlinbert2019}
Jacob Devlin, Ming{-}Wei Chang, Kenton Lee, and Kristina Toutanova. 2019.
\newblock \href {https://doi.org/10.18653/v1/n19-1423} {{BERT:} pre-training of
  deep bidirectional transformers for language understanding}.
\newblock In \emph{Proceedings of the 2019 Conference of the North American
  Chapter of the Association for Computational Linguistics: Human Language
  Technologies, {NAACL-HLT} 2019, Minneapolis, MN, USA, June 2-7, 2019, Volume
  1 (Long and Short Papers)}, pages 4171--4186. Association for Computational
  Linguistics.

\bibitem[{Fedus et~al.(2021)Fedus, Zoph, and Shazeer}]{fedus2021switch}
William Fedus, Barret Zoph, and Noam Shazeer. 2021.
\newblock Switch transformers: Scaling to trillion parameter models with simple
  and efficient sparsity.
\newblock \emph{arXiv preprint arXiv:2101.03961}.

\bibitem[{Google(2020)}]{xprof}
Google. 2020.
\newblock {P}rofile your model with cloud tpu tools.
\newblock \url{https://cloud.google.com/tpu/docs/cloud-tpu-tools}.
\newblock Accessed: 2022-11-11.

\bibitem[{Guo et~al.(2022)Guo, Ainslie, Uthus, Onta\~{n}\'{o}n, Ni, Sung, and
  Yang}]{guo2022longt5}
Mandy Guo, Joshua Ainslie, David Uthus, Santiago Onta\~{n}\'{o}n, Jianmo Ni,
  Yun-Hsuan Sung, and Yinfei Yang. 2022.
\newblock \href {https://doi.org/10.18653/v1/2022.findings-naacl.55}
  {{L}ong{T}5: {E}fficient text-to-text transformer for long sequences}.
\newblock In \emph{Findings of the Association for Computational Linguistics:
  NAACL 2022}, pages 724--736, Seattle, United States. Association for
  Computational Linguistics.

\bibitem[{Heek et~al.(2020)Heek, Levskaya, Oliver, Ritter, Rondepierre,
  Steiner, and van {Z}ee}]{flax}
Jonathan Heek, Anselm Levskaya, Avital Oliver, Marvin Ritter, Bertrand
  Rondepierre, Andreas Steiner, and Marc van {Z}ee. 2020.
\newblock \href {http://github.com/google/flax} {{F}lax: A neural network
  library and ecosystem for {JAX}}.

\bibitem[{Huang et~al.(2021)Huang, Cao, Parulian, Ji, and
  Wang}]{huang2021govreport}
Luyang Huang, Shuyang Cao, Nikolaus Parulian, Heng Ji, and Lu~Wang. 2021.
\newblock \href {https://doi.org/10.18653/v1/2021.naacl-main.112} {Efficient
  attentions for long document summarization}.
\newblock In \emph{Proceedings of the 2021 Conference of the North American
  Chapter of the Association for Computational Linguistics: Human Language
  Technologies}, pages 1419--1436, Online. Association for Computational
  Linguistics.

\bibitem[{Joshi et~al.(2017)Joshi, Choi, Weld, and
  Zettlemoyer}]{joshi2017triviaqa}
Mandar Joshi, Eunsol Choi, Daniel~S. Weld, and Luke Zettlemoyer. 2017.
\newblock Triviaqa: A large scale distantly supervised challenge dataset for
  reading comprehension.
\newblock In \emph{Proceedings of the 55th Annual Meeting of the Association
  for Computational Linguistics}, Vancouver, Canada. Association for
  Computational Linguistics.

\bibitem[{Kaplan et~al.(2020)Kaplan, McCandlish, Henighan, Brown, Chess, Child,
  Gray, Radford, Wu, and Amodei}]{kaplan2020scaling}
Jared Kaplan, Sam McCandlish, Tom Henighan, Tom~B. Brown, Benjamin Chess, Rewon
  Child, Scott Gray, Alec Radford, Jeffrey Wu, and Dario Amodei. 2020.
\newblock \href {http://arxiv.org/abs/2001.08361} {Scaling laws for neural
  language models}.
\newblock \emph{CoRR}, abs/2001.08361.

\bibitem[{Ko{\v{c}}isk{\'y} et~al.(2018)Ko{\v{c}}isk{\'y}, Schwarz, Blunsom,
  Dyer, Hermann, Melis, and Grefenstette}]{kocisky2018narrativeqa}
Tom{\'a}{\v{s}} Ko{\v{c}}isk{\'y}, Jonathan Schwarz, Phil Blunsom, Chris Dyer,
  Karl~Moritz Hermann, G{\'a}bor Melis, and Edward Grefenstette. 2018.
\newblock \href {https://doi.org/10.1162/tacl_a_00023} {The {N}arrative{QA}
  reading comprehension challenge}.
\newblock \emph{Transactions of the Association for Computational Linguistics},
  6:317--328.

\bibitem[{Koreeda and Manning(2021)}]{koreeda-manning-2021-contractnli}
Yuta Koreeda and Christopher Manning. 2021.
\newblock \href {https://doi.org/10.18653/v1/2021.findings-emnlp.164}
  {{C}ontract{NLI}: A dataset for document-level natural language inference for
  contracts}.
\newblock In \emph{Findings of the Association for Computational Linguistics:
  EMNLP 2021}, pages 1907--1919, Punta Cana, Dominican Republic. Association
  for Computational Linguistics.

\bibitem[{Kratzwald and Feuerriegel(2018)}]{adaptiveretrieval}
Bernhard Kratzwald and Stefan Feuerriegel. 2018.
\newblock \href {https://doi.org/10.18653/v1/d18-1055} {Adaptive document
  retrieval for deep question answering}.
\newblock In \emph{Proceedings of the 2018 Conference on Empirical Methods in
  Natural Language Processing, Brussels, Belgium, October 31 - November 4,
  2018}, pages 576--581. Association for Computational Linguistics.

\bibitem[{Kwiatkowski et~al.(2019)Kwiatkowski, Palomaki, Redfield, Collins,
  Parikh, Alberti, Epstein, Polosukhin, Devlin, Lee, Toutanova, Jones, Kelcey,
  Chang, Dai, Uszkoreit, Le, and Petrov}]{kiwatkowski2019nq}
Tom Kwiatkowski, Jennimaria Palomaki, Olivia Redfield, Michael Collins,
  Ankur~P. Parikh, Chris Alberti, Danielle Epstein, Illia Polosukhin, Jacob
  Devlin, Kenton Lee, Kristina Toutanova, Llion Jones, Matthew Kelcey,
  Ming{-}Wei Chang, Andrew~M. Dai, Jakob Uszkoreit, Quoc Le, and Slav Petrov.
  2019.
\newblock \href {https://doi.org/10.1162/tacl\_a\_00276} {Natural questions: a
  benchmark for question answering research}.
\newblock \emph{Trans. Assoc. Comput. Linguistics}, 7:452--466.

\bibitem[{Lei et~al.(2023)Lei, Bai, Brahma, Ainslie, Lee, Zhou, Du, Zhao, Wu,
  Li, Zhang, and Chang}]{lei2023conditional}
Tao Lei, Junwen Bai, Siddhartha Brahma, Joshua Ainslie, Kenton Lee, Yanqi Zhou,
  Nan Du, Vincent~Y. Zhao, Yuexin Wu, Bo~Li, Yu~Zhang, and Ming-Wei Chang.
  2023.
\newblock Conditional adapters: Parameter-efficient transfer learning with fast
  inference.
\newblock In \emph{Advances in Neural Information Processing Systems}.

\bibitem[{Mao et~al.(2021)Mao, He, Liu, Shen, Gao, Han, and
  Chen}]{readerguidererank}
Yuning Mao, Pengcheng He, Xiaodong Liu, Yelong Shen, Jianfeng Gao, Jiawei Han,
  and Weizhu Chen. 2021.
\newblock \href {https://doi.org/10.18653/v1/2021.findings-acl.29}
  {Reader-guided passage reranking for open-domain question answering}.
\newblock In \emph{Findings of the Association for Computational Linguistics:
  {ACL/IJCNLP} 2021, Online Event, August 1-6, 2021}, volume {ACL/IJCNLP} 2021
  of \emph{Findings of {ACL}}, pages 344--350. Association for Computational
  Linguistics.

\bibitem[{Pang et~al.(2021)Pang, Parrish, Joshi, Nangia, Phang, Chen,
  Padmakumar, Ma, Thompson, He, and Bowman}]{pang2021quality}
Richard~Yuanzhe Pang, Alicia Parrish, Nitish Joshi, Nikita Nangia, Jason Phang,
  Angelica Chen, Vishakh Padmakumar, Johnny Ma, Jana Thompson, He~He, and
  Samuel~R. Bowman. 2021.
\newblock {QuALITY}: Question answering with long input texts, yes!
\newblock \emph{arXiv preprint arXiv:2112.08608}.

\bibitem[{Pope et~al.(2022)Pope, Douglas, Chowdhery, Devlin, Bradbury,
  Levskaya, Heek, Xiao, Agrawal, and Dean}]{pope2022efficiently}
Reiner Pope, Sholto Douglas, Aakanksha Chowdhery, Jacob Devlin, James Bradbury,
  Anselm Levskaya, Jonathan Heek, Kefan Xiao, Shivani Agrawal, and Jeff Dean.
  2022.
\newblock Efficiently scaling transformer inference.
\newblock \emph{arXiv preprint arXiv:2211.05102}.

\bibitem[{Qian et~al.(2022)Qian, Lee, Duddu, Dai, Brahma, Naim, Lei, and
  Zhao}]{qian2022multi}
Yujie Qian, Jinhyuk Lee, Sai Meher~Karthik Duddu, Zhuyun Dai, Siddhartha
  Brahma, Iftekhar Naim, Tao Lei, and Vincent~Y Zhao. 2022.
\newblock \href {https://arxiv.org/abs/2211.01267} {Multi-vector retrieval as
  sparse alignment}.
\newblock \emph{arXiv preprint arXiv:2211.01267}.

\bibitem[{Raffel et~al.(2020)Raffel, Shazeer, Roberts, Lee, Narang, Matena,
  Zhou, Li, and Liu}]{raffel2020t5}
Colin Raffel, Noam Shazeer, Adam Roberts, Katherine Lee, Sharan Narang, Michael
  Matena, Yanqi Zhou, Wei Li, and Peter~J. Liu. 2020.
\newblock \href {http://jmlr.org/papers/v21/20-074.html} {Exploring the limits
  of transfer learning with a unified text-to-text transformer}.
\newblock \emph{J. Mach. Learn. Res.}, 21:140:1--140:67.

\bibitem[{Roberts et~al.(2022)Roberts, Chung, Levskaya, Mishra, Bradbury,
  Andor, Narang, Lester, Gaffney, Mohiuddin, Hawthorne, Lewkowycz, Salcianu,
  van Zee, Austin, Goodman, Soares, Hu, Tsvyashchenko, Chowdhery, Bastings,
  Bulian, Garcia, Ni, Chen, Kenealy, Clark, Lee, Garrette, Lee-Thorp, Raffel,
  Shazeer, Ritter, Bosma, Passos, Maitin-Shepard, Fiedel, Omernick, Saeta,
  Sepassi, Spiridonov, Newlan, and Gesmundo}]{t5x}
Adam Roberts, Hyung~Won Chung, Anselm Levskaya, Gaurav Mishra, James Bradbury,
  Daniel Andor, Sharan Narang, Brian Lester, Colin Gaffney, Afroz Mohiuddin,
  Curtis Hawthorne, Aitor Lewkowycz, Alex Salcianu, Marc van Zee, Jacob Austin,
  Sebastian Goodman, Livio~Baldini Soares, Haitang Hu, Sasha Tsvyashchenko,
  Aakanksha Chowdhery, Jasmijn Bastings, Jannis Bulian, Xavier Garcia, Jianmo
  Ni, Andrew Chen, Kathleen Kenealy, Jonathan~H. Clark, Stephan Lee, Dan
  Garrette, James Lee-Thorp, Colin Raffel, Noam Shazeer, Marvin Ritter, Maarten
  Bosma, Alexandre Passos, Jeremy Maitin-Shepard, Noah Fiedel, Mark Omernick,
  Brennan Saeta, Ryan Sepassi, Alexander Spiridonov, Joshua Newlan, and Andrea
  Gesmundo. 2022.
\newblock \href {https://arxiv.org/abs/2203.17189} {Scaling up models and data
  with $\texttt{t5x}$ and $\texttt{seqio}$}.
\newblock \emph{arXiv preprint arXiv:2203.17189}.

\bibitem[{Schuster et~al.(2022)Schuster, Fisch, Gupta, Dehghani, Bahri, Tran,
  Tay, and Metzler}]{schuster2022confident}
Tal Schuster, Adam Fisch, Jai Gupta, Mostafa Dehghani, Dara Bahri, Vinh~Q Tran,
  Yi~Tay, and Donald Metzler. 2022.
\newblock Confident adaptive language modeling.
\newblock \emph{arXiv preprint arXiv:2207.07061}.

\bibitem[{Shaham et~al.(2022)Shaham, Segal, Ivgi, Efrat, Yoran, Haviv, Gupta,
  Xiong, Geva, Berant, and Levy}]{shaham2022scrolls}
Uri Shaham, Elad Segal, Maor Ivgi, Avia Efrat, Ori Yoran, Adi Haviv, Ankit
  Gupta, Wenhan Xiong, Mor Geva, Jonathan Berant, and Omer Levy. 2022.
\newblock Scrolls: Standardized comparison over long language sequences.
\newblock \emph{ArXiv}, abs/2201.03533.

\bibitem[{Shazeer(2019)}]{shazeer2019mq}
Noam Shazeer. 2019.
\newblock Fast transformer decoding: One write-head is all you need.
\newblock \emph{arXiv preprint arXiv:1911.02150}.

\bibitem[{Shazeer et~al.(2017)Shazeer, Mirhoseini, Maziarz, Davis, Le, Hinton,
  and Dean}]{shazeer2017moe}
Noam Shazeer, Azalia Mirhoseini, Krzysztof Maziarz, Andy Davis, Quoc~V. Le,
  Geoffrey~E. Hinton, and Jeff Dean. 2017.
\newblock \href {https://openreview.net/forum?id=B1ckMDqlg} {Outrageously large
  neural networks: The sparsely-gated mixture-of-experts layer}.
\newblock In \emph{5th International Conference on Learning Representations,
  {ICLR} 2017, Toulon, France, April 24-26, 2017, Conference Track
  Proceedings}. OpenReview.net.

\bibitem[{Shazeer and Stern(2018)}]{adafactor}
Noam Shazeer and Mitchell Stern. 2018.
\newblock \href {http://proceedings.mlr.press/v80/shazeer18a.html} {Adafactor:
  Adaptive learning rates with sublinear memory cost}.
\newblock In \emph{Proceedings of the 35th International Conference on Machine
  Learning, {ICML} 2018, Stockholmsm{\"{a}}ssan, Stockholm, Sweden, July 10-15,
  2018}, volume~80 of \emph{Proceedings of Machine Learning Research}, pages
  4603--4611. {PMLR}.

\bibitem[{Tay et~al.(2021)Tay, Dehghani, Abnar, Shen, Bahri, Pham, Rao, Yang,
  Ruder, and Metzler}]{tay2021long}
Yi~Tay, Mostafa Dehghani, Samira Abnar, Yikang Shen, Dara Bahri, Philip Pham,
  Jinfeng Rao, Liu Yang, Sebastian Ruder, and Donald Metzler. 2021.
\newblock \href {https://openreview.net/forum?id=qVyeW-grC2k} {Long range arena
  : A benchmark for efficient transformers}.
\newblock In \emph{International Conference on Learning Representations}.

\bibitem[{Tay et~al.(2022)Tay, Dehghani, Tran, Garcia, Bahri, Schuster, Zheng,
  Houlsby, and Metzler}]{tay2022ul2}
Yi~Tay, Mostafa Dehghani, Vinh~Q Tran, Xavier Garcia, Dara Bahri, Tal Schuster,
  Huaixiu~Steven Zheng, Neil Houlsby, and Donald Metzler. 2022.
\newblock Unifying language learning paradigms.
\newblock \emph{arXiv preprint arXiv:2205.05131}.

\bibitem[{Varshney et~al.(2022)Varshney, Luo, and Baral}]{canext}
Neeraj Varshney, Man Luo, and Chitta Baral. 2022.
\newblock \href {https://doi.org/10.48550/arXiv.2211.12707} {Can open-domain
  {QA} reader utilize external knowledge efficiently like humans?}
\newblock \emph{CoRR}, abs/2211.12707.

\bibitem[{Wang et~al.(2018)Wang, Yu, Guo, Wang, Klinger, Zhang, Chang, Tesauro,
  Zhou, and Jiang}]{r3rerank}
Shuohang Wang, Mo~Yu, Xiaoxiao Guo, Zhiguo Wang, Tim Klinger, Wei Zhang, Shiyu
  Chang, Gerry Tesauro, Bowen Zhou, and Jing Jiang. 2018.
\newblock \href
  {https://www.aaai.org/ocs/index.php/AAAI/AAAI18/paper/view/16712}
  {R\({}^{\mbox{3}}\): Reinforced ranker-reader for open-domain question
  answering}.
\newblock In \emph{Proceedings of the Thirty-Second {AAAI} Conference on
  Artificial Intelligence, (AAAI-18), the 30th innovative Applications of
  Artificial Intelligence (IAAI-18), and the 8th {AAAI} Symposium on
  Educational Advances in Artificial Intelligence (EAAI-18), New Orleans,
  Louisiana, USA, February 2-7, 2018}, pages 5981--5988. {AAAI} Press.

\bibitem[{Wang et~al.(2020)Wang, Li, Khabsa, Fang, and Ma}]{wang2020linformer}
Sinong Wang, Belinda~Z Li, Madian Khabsa, Han Fang, and Hao Ma. 2020.
\newblock Linformer: Self-attention with linear complexity.
\newblock \emph{arXiv preprint arXiv:2006.04768}.

\bibitem[{Yu et~al.(2022)Yu, Zhu, Fang, Yu, Wang, Xu, Ren, Yang, and
  Zeng}]{kgfid}
Donghan Yu, Chenguang Zhu, Yuwei Fang, Wenhao Yu, Shuohang Wang, Yichong Xu,
  Xiang Ren, Yiming Yang, and Michael Zeng. 2022.
\newblock \href {https://doi.org/10.18653/v1/2022.acl-long.340} {Kg-fid:
  Infusing knowledge graph in fusion-in-decoder for open-domain question
  answering}.
\newblock In \emph{Proceedings of the 60th Annual Meeting of the Association
  for Computational Linguistics (Volume 1: Long Papers), {ACL} 2022, Dublin,
  Ireland, May 22-27, 2022}, pages 4961--4974. Association for Computational
  Linguistics.

\bibitem[{Zaheer et~al.(2020)Zaheer, Guruganesh, Dubey, Ainslie, Alberti,
  Onta\~{n}\'{o}n, Pham, Ravula, Wang, Yang et~al.}]{zaheer2020big}
Manzil Zaheer, Guru Guruganesh, Kumar~Avinava Dubey, Joshua Ainslie, Chris
  Alberti, Santiago Onta\~{n}\'{o}n, Philip Pham, Anirudh Ravula, Qifan Wang,
  Li~Yang, et~al. 2020.
\newblock Big bird: Transformers for longer sequences.
\newblock \emph{Advances in Neural Information Processing Systems},
  33:17283--17297.

\bibitem[{Zemlyanskiy et~al.(2021)Zemlyanskiy, Ainslie, de~Jong, Pham,
  Eckstein, and Sha}]{zemlyanskiy2021readtwice}
Yury Zemlyanskiy, Joshua Ainslie, Michiel de~Jong, Philip Pham, Ilya Eckstein,
  and Fei Sha. 2021.
\newblock Readtwice: Reading very large documents with memories.
\newblock In \emph{Proceedings of the 2021 Conference of the North American
  Chapter of the Association for Computational Linguistics: Human Language
  Technologies}, pages 5189--5195.

\bibitem[{Zhang et~al.(2020)Zhang, Zhao, Saleh, and Liu}]{zhang2020pegasus}
Jingqing Zhang, Yao Zhao, Mohammad Saleh, and Peter Liu. 2020.
\newblock Pegasus: Pre-training with extracted gap-sentences for abstractive
  summarization.
\newblock In \emph{International Conference on Machine Learning}, pages
  11328--11339. PMLR.

\bibitem[{Zhong et~al.(2021)Zhong, Yin, Yu, Zaidi, Mutuma, Jha, Awadallah,
  Celikyilmaz, Liu, Qiu, and Radev}]{zhong2021qmsum}
Ming Zhong, Da~Yin, Tao Yu, Ahmad Zaidi, Mutethia Mutuma, Rahul Jha,
  Ahmed~Hassan Awadallah, Asli Celikyilmaz, Yang Liu, Xipeng Qiu, and Dragomir
  Radev. 2021.
\newblock \href {https://doi.org/10.18653/v1/2021.naacl-main.472} {{QMS}um: A
  new benchmark for query-based multi-domain meeting summarization}.
\newblock In \emph{Proceedings of the 2021 Conference of the North American
  Chapter of the Association for Computational Linguistics: Human Language
  Technologies}, pages 5905--5921, Online. Association for Computational
  Linguistics.

\bibitem[{Zoph et~al.(2022)Zoph, Bello, Kumar, Du, Huang, Dean, Shazeer, and
  Fedus}]{zoph2022stmoe}
Barret Zoph, Irwan Bello, Sameer Kumar, Nan Du, Yanping Huang, Jeff Dean, Noam
  Shazeer, and William Fedus. 2022.
\newblock St-moe: Designing stable and transferable sparse expert models.
\newblock \emph{arXiv preprint arXiv:2202.08906}.

\end{thebibliography}
\bibliographystyle{acl_natbib}

\clearpage
\begin{table*}[t!]
\small
\centering
\vspace{0.35cm}
\begin{tabular}{l|ccccccc}
    \textbf{Model} & {\bf Layers} & {\bf Model dim} & {\bf MLP\textsubscript{light} dim} & {\bf MLP\textsubscript{heavy} dim} & {\bf Heads\textsubscript{light}} & {\bf Heads\textsubscript{heavy}} & {\bf Params}  \\ 
    \toprule 
    \longt-B & 12 & 768 & 2048 & N/A &  12 & N/A & 248m \\
    \slic-B & 12 & 768 & 1024 & 8096 &  4 & 8 & 433m \\
    \midrule
    \longt-L & 24 & 1024 & 2816 & N/A &  16 & N/A & 783m \\
    \slic-L & 24 & 1024 & 1408 & 11264 &  4 & 12 & 1462m \\
    \midrule
    \longt-XL & 24 & 2048 & 5120 & N/A &  32 & N/A & 2850m \\
    \slic-XL & 24 & 2048 & 2560 & 20480 &  8 & 24 & 5297m \\    
    \bottomrule
\end{tabular}
\caption{Hyperparameters for \longt and \slic models. T5.1.1 hyperparameters match \longt. \slic parameters are sparsely accessed as a result of conditional computation, so parameter counts do not reflect compute, and for a given model size \slic is in fact faster than \longt despite having more parameters.}
\label{table:parameters}
\end{table*}

\begin{table*}[b]
\centering
\small
\begin{tabular}{l|cc|cc|cc|cc|cc}
\toprule
{\bf Model} & \multicolumn{2}{c|}{\bf Average} & \multicolumn{2}{c|}{\bf 16k in, 128 out} & \multicolumn{2}{c|}{\bf 16k in, 512 out} &
\multicolumn{2}{c|}{\bf 16k in, 1024 out} & \multicolumn{2}{c}{\bf 8k in, 128 out}\\
\midrule
&{\bf Enc} & {\bf Tot} &{\bf Enc} & {\bf Tot} &{\bf Enc} & {\bf Tot} &{\bf Enc} & {\bf Tot} &{\bf Enc} & {\bf Tot}  \\
\midrule
\longt-B& 77 & 136 & 84 & 98 & 84 & 165 & 84 &  296 & 27 & 39 \\
\slic-B& 29 & 90 & 30 & 45 & 30 & 113 & 30 & 256 & 18 & 30  \\
\midrule
\longt-L& 164 & 329 & 173 & 222 & 179 & 392 & 179 & 799 &  66 & 100 \\
\slic-L & 70 & 201 & 73 & 103 & 73 & 250 & 73 & 578 & 45 & 69 \\
\midrule
\longt-XL & 390 & 870 & 412 & 557 & 423 & 1081 & 423 & 2065 & 166 & 290 \\
\slic-XL & 177 & 439 & 185 & 239 & 185 & 525 & 185 & 1253 & 115 & 163\\
\bottomrule
\end{tabular}
\caption{Comparison of total and encoder inference time per sample (ms) for \longt and \slic Base, Large, and XL models at different input and output lengths. Average time per sample is computed as a weighted average over input and output lengths, weighted by the number of tasks in our evaluation that use the corresponding setting (4 for 16k/128, 3 for 16k/512, and one each for 16k/1024 and 8k/128).}
\label{table:full_timing}
\end{table*}
\appendix

\section{Contributions}
\label{sec:appendix-contributions}
Joshua led the project, developed the initial conditional attention mechanisms, and conducted most experimental ablations.  Tao developed the heavy/light formulation for heterogeneous conditional computation, comprising the routing and conditional feedforward mechanisms, and iterated with Joshua on initial experiments demonstrating feasibility.  Michiel helped to scope the paper, performed most of the writing, and oversaw speed benchmarking.  Santiago designed and conducted all the few-shot experiments, initiated the routing analysis visualization, and integrated UL2 into the codebase.  Siddhartha developed the separate routing for query and key/value tokens in the conditional attention component and demonstrated the resulting quality improvements.  Yury designed and conducted all experiments for inputs larger than 16k tokens, demonstrating favorable scaling up to 64k.  David integrated all SCROLLS tasks into the codebase and ran early experiments, especially comparing UL2 with PEGASUS.  Mandy developed the leaderboard comparisons with LongT5 and helped run several experiments.  James advised on and ran early comparisons with MoE conditional computation.  Yi advised on the adaptation of UL2 to 4k input length pre-training.  Finally, Yun-Hsuan and Sumit provided guidance and support for the project overall.

\section{Model Hyperparameters}
\label{sec:appendix-parameters}

Table \ref{table:parameters} shows \longt and \slic hyperparameters, including parameter counts. For \longt, we report numbers for the TGlobal configuration, which match T5.1.1. Notice that \slic's parameter counts are larger due to using conditional compute. Similar to other conditional compute architectures such as mixture-of-experts, computational cost does not necessarily increase with parameter count.

We use the same 127-token local radius for \slic as \longt.  This results in a local attention window $w$ of 255 since 127 tokens are attended to the left and 127 to the right.

\section{Routing Normalization Hyperparameters}
\label{sec:appendix-routing-hyperparams}

To normalize the routing scores for differentiable top-$k$ token selection, we use the iterative soft top-$k$ algorithm from \citet{lei2023conditional} and \citet{qian2022multi} with $\epsilon = 1.0$ and 50 iterations.  During training we allow the top $\frac{9}{8} k$ tokens to have nonzero weight instead of just the top $k$ in order to provide a slightly improved training signal.

\section{Additional Experimental Results}
\label{sec:appendix-results}

Table \ref{table:full_timing} compares \longt and \slic inference speed in more detail, splitting off encoder and total time per sample. Since \slic applies conditional computation only in the encoder, encoder speed gains are larger than overall speed gain, and total speed gains are largest for shorter output length. Trade-offs are even more in the favor of \slic when paired with other decoder optimizations.

\begin{table*}[t]

\centering
\small
\begin{tabular}{l|ccc|ccc|ccc|ccc}
\toprule
\textbf{Model}  & \multicolumn{3}{c|}{\textbf{arXiv}}& \multicolumn{3}{c|}{\textbf{SummScreenFD}} & \multicolumn{3}{c|}{\textbf{QMSum}} & \multicolumn{3}{c}{\textbf{GovRep}}    \\
\midrule
& \textbf{R-1} & \textbf{R-2} & \textbf{R-L} & \textbf{R-1} & \textbf{R-2} & \textbf{R-L} & \textbf{R-1} & \textbf{R-2} & \textbf{R-L} & \textbf{R-1} & \textbf{R-2} & \textbf{R-L}\\
\midrule
 \longt-B & 47.4 & 21.4 & 43.5 & 34.8 & 9.3 & 20.7 & 35.1 & 11.1 &   23.4 & 59.3 & 30.1 & 33.0  \\    
\slic-B & 47.5  & 21.3 & 43.6 &  35.6 & 9.7 & 21.0 & 34.6 & 10.9 & 23.0 & 60.2 & 31.0 & 32.8 \\
\midrule
\longt-L  & 47.9 & 21.7 & 43.8 & 35.3 & 9.1 & 20.8 & 35.9 & 12.0 & 24.1 & 61.4 & 32.5 & 34.1 \\    
 \slic-L  & 48.4 & 21.7 & 44.3 & 35.7 & 10.1 & 21.4 & 36.8 & 12.6 & 24.7 & 61.8 & 32.7 & 34.4 \\ 
\midrule
\longt-XL  & 48.2 & 21.8 & 44.1 &  36.6 & 10.3 & 21.5 & 37.0 & 12.5  & 24.7 & 61.8 & 33.2 & 34.8\\    
 \slic-XL & 48.4 & 22.0 & 44.3 &  36.3 & 10.0 & 21.5 & 37.4 & 13.0 & 25.1 & 62.2 & 33.3 & 34.9\\  
    \bottomrule
\end{tabular}
\caption{Full performance comparison with Rouge-1, Rouge-2, and Rouge-L metrics of \slic and \longt Base, Large, and XL models on summarization dev sets. Results based on checkpoint that maximizes R\textsubscript{gm} as in Table \ref{table:headline_results}.}
\label{table:full_summarization_results}

\end{table*}

Table \ref{table:full_summarization_results} shows full (Rouge-1, Rouge-2, Rouge-L) results for summarization datasets.

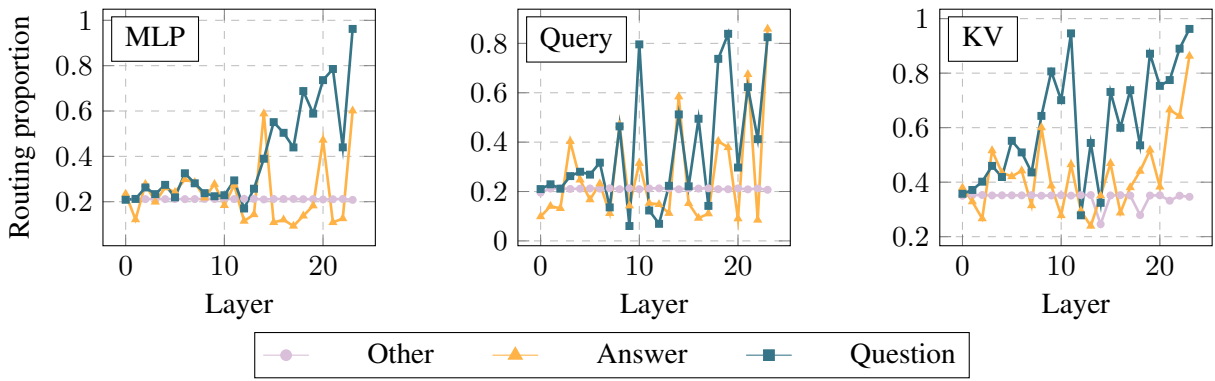
\begin{figure*}[b]
     \centering
     \begin{subfigure}[t]{0.28\textwidth}
        \raggedright
        \begin{tikzpicture}[scale=1.0]
            \begin{axis}[
            scale only axis,
            width=0.8\textwidth,
            height=0.7\textwidth,
            xlabel={Layer},
            ylabel={Routing proportion},
            mark=x,
            x tick label style={log ticks with fixed point},
            ymajorgrids=true,
            xmajorgrids=true,
            xminorticks=true,
            grid style=dashed,
            legend pos={north west},
            ylabel style={align=left, text height=0.2cm},
            ]
            \addlegendimage{empty legend}\addlegendentry{MLP}
            \addplot[color=othercolor,mark=\othermark,mark size=1pt, line width=1] table {
            0 0.21161997946429376
            1 0.2125872223817629
            2 0.21139287510891192
            3 0.2120075226278527
            4 0.21143393511373654
            5 0.21185861990537036
            6 0.2113462075224012
            7 0.21165094318084943
            8 0.21246731457213297
            9 0.21207230297236887
            10 0.21240816743978747
            11 0.21173454768636082
            12 0.21308165538967852
            13 0.21261318091433634
            14 0.20987843682723828
            15 0.2119161537179608
            16 0.21203253878894812
            17 0.21229383346521144
            18 0.21107189040051932
            19 0.21156451254256683
            20 0.20975093698681574
            21 0.21084240583540229
            22 0.21221006045407584
            23 0.2080541786840686
            };
            \addplot[color=answercolor,mark=\answermark,mark size=1pt, line width=1] table {
            0 0.23404124981875588
            1 0.12133974985640475
            2 0.2773626670220826
            3 0.1990351298117969
            4 0.26216721309024876
            5 0.24163464687057307
            6 0.2982911783714974
            7 0.2850873355229606
            8 0.21702661150057342
            9 0.27718464853413505
            10 0.18387851142672684
            11 0.2737455132017648
            12 0.11462717450160506
            13 0.14369442677229646
            14 0.5881649708680756
            15 0.10881932430568753
            16 0.12051417298406941
            17 0.09281284504910445
            18 0.1376113019123887
            19 0.18240325725650058
            20 0.4718293490780316
            21 0.10874523946867246
            22 0.12584635703993513
            23 0.6010882598061337
            };
            \addplot[color=questioncolor,mark=\questionmark,mark size=1pt,line width=1] table {
            0 0.20856889202407008
            1 0.21226976756629298
            2 0.2629011370564126
            3 0.2335030021419846
            4 0.2742826105589716
            5 0.21929366584665527
            6 0.325394583890978
            7 0.28125829559056403
            8 0.23776303913582714
            9 0.22415379958167925
            10 0.2285775261957093
            11 0.2940282392779054
            12 0.1701509477917461
            13 0.25734053143717994
            14 0.3895200056455131
            15 0.5509839725342021
            16 0.5038410777931918
            17 0.43938062321848653
            18 0.6880240312117615
            19 0.5884735326060652
            20 0.7361686221028755
            21 0.7852749420317643
            22 0.43978855486954743
            23 0.9620789504443227
            };               
        
            \end{axis}       
        \end{tikzpicture}  
     \end{subfigure} \hspace{0.36cm}
     \hfill
     \begin{subfigure}[t]{0.28\textwidth}
        \centering
        \begin{tikzpicture}[scale=1.0]
            \begin{axis}[
            scale only axis,
            width=0.8\textwidth,
            height=0.7\textwidth,
            xlabel={Layer},
            mark=x,
            x tick label style={log ticks with fixed point},
            ymajorgrids=true,
            xmajorgrids=true,
            xminorticks=true,
            grid style=dashed,
            legend pos={north west},
            ylabel style={align=left, text height=0.2cm},
            ]
            \addlegendimage{empty legend}\addlegendentry{Query}
            \addplot[color=othercolor,mark=\othermark,mark size=1pt, line width=1] table {
            0 0.19549139457384002
            1 0.21209221538348624
            2 0.2125791879007595
            3 0.2105169686202275
            4 0.2112944328572769
            5 0.21240783221189677
            6 0.21176481164882838
            7 0.2130972302762573
            8 0.20935626333610893
            9 0.2130512618262021
            10 0.20952047660539894
            11 0.21263229572024034
            12 0.21287331516477234
            13 0.2125744576897517
            14 0.20913046305089916
            15 0.21240088471819554
            16 0.21199353612183974
            17 0.21297168167351868
            18 0.20954572724295997
            19 0.20935358805721618
            20 0.21268238211396587
            21 0.20873566927429005
            22 0.21232251657333964
            23 0.20697159425681427
            };   
            \addplot[color=answercolor,mark=\answermark,mark size=1pt, line width=1] table {
            0 0.09856966095334675
            1 0.14060753632977327
            2 0.13186202080930925
            3 0.40304072073742486
            4 0.24739154421037465
            5 0.1672355452586664
            6 0.23150961204544812
            7 0.11160585670330919
            8 0.46910581322547396
            9 0.14235919997308963
            10 0.3146520498904203
            11 0.15289618580039627
            12 0.1479256285183321
            13 0.11205143872629633
            14 0.5839836673398893
            15 0.1519279027796636
            16 0.09239308751278266
            17 0.11036272263682104
            18 0.4037651750456416
            19 0.37856251427068704
            20 0.0901048019000607
            21 0.6743836892905413
            22 0.08463163214863184
            23 0.8586354041149576
            };               
            \addplot[color=questioncolor,mark=\questionmark,mark size=1pt, line width=1] table {
            0 0.2099036197909046
            1 0.22926699697975958
            2 0.211451565719695
            3 0.26243562420203304
            4 0.2800879162310573
            5 0.2684250004017433
            6 0.3169072893202554
            7 0.13571953945826448
            8 0.463885009902794
            9 0.06026191930443222
            10 0.796024658192565
            11 0.12316350899062913
            12 0.0691505313549957
            13 0.22311366730813595
            14 0.5122468448194271
            15 0.22110591251264508
            16 0.49468819575532663
            17 0.14179612527010294
            18 0.7373704741411858
            19 0.8387805202703116
            20 0.2971193860421291
            21 0.6228262860671179
            22 0.412059198614116
            23 0.8251401341248926
            };             
            \end{axis}       
        \end{tikzpicture}  
     \end{subfigure}     
     \hfill
     \begin{subfigure}[t]{0.28\textwidth}
        \raggedleft
        \begin{tikzpicture}[scale=1.0]
            \begin{axis}[
            scale only axis,
            width=0.8\textwidth,
            height=0.7\textwidth,
            xlabel={Layer},
            mark=x,
            x tick label style={log ticks with fixed point},
            ymajorgrids=true,
            xmajorgrids=true,
            xminorticks=true,
            grid style=dashed,
            legend pos={north west},
            ylabel style={align=left, text height=0.2cm},
            ]
            \addlegendimage{empty legend}\addlegendentry{KV}
            \addplot[color=othercolor,mark=\othermark,mark size=1pt, line width=1] table {
            0 0.35019249035137523
            1 0.3525438069145887
            2 0.35293969478952314
            3 0.3512627081448169
            4 0.35180610639939575
            5 0.3516696604571174
            6 0.3516573057818594
            7 0.35250147529909165
            8 0.3502812027267641
            9 0.35110776222753687
            10 0.3519654084233534
            11 0.35073425660566815
            12 0.3526173688518903
            13 0.3527463777811796
            14 0.2456468071859964
            15 0.35132045050321287
            16 0.3525094426405284
            17 0.3510333293545492
            18 0.2792762573520494
            19 0.35100672693816265
            20 0.35176224045963533
            21 0.3321378796247362
            22 0.3503188617750819
            23 0.34622569438400463
            };   
            \addplot[color=answercolor,mark=\answermark,mark size=1pt, line width=1] table {
            0 0.37919229692764433
            1 0.32862221726090235
            2 0.2663435888759715
            3 0.5158641264872711
            4 0.4348650637532726
            5 0.42079635832020107
            6 0.44208448995530036
            7 0.31435037054317644
            8 0.6003177818950836
            9 0.3867071926342268
            10 0.27786296098789925
            11 0.464732708828246
            12 0.29523005085614473
            13 0.2391429766550423
            14 0.35034875120751774
            15 0.4691989400084632
            16 0.287680735209191
            17 0.3805258248305361
            18 0.4406435230754643
            19 0.5181929256800606
            20 0.38277884263242634
            21 0.6655974345329255
            22 0.6420381393763785
            23 0.862727037011765
            };               
            \addplot[color=questioncolor,mark=\questionmark,mark size=1pt, line width=1] table {
            0 0.3574036293389754
            1 0.3714281853222321
            2 0.40176561417062046
            3 0.45976470574889095
            4 0.41888616643100607
            5 0.5517346042962912
            6 0.5092366638237913
            7 0.4357807900764302
            8 0.6431377355400922
            9 0.8062974080445435
            10 0.7008200519289048
            11 0.9459240178802629
            12 0.2784017532337479
            13 0.5438503167829559
            14 0.32421232699312585
            15 0.7310340349223263
            16 0.59942834172575
            17 0.7376625469667542
            18 0.5351831860687264
            19 0.8712681993442024
            20 0.7532590154318346
            21 0.7746499807418749
            22 0.8897478233991253
            23 0.962236743816725
            };            
            \end{axis}    
        \end{tikzpicture}  
     \end{subfigure}   
     \hfill
     
     \begin{tikzpicture}
        \begin{customlegend}[
            legend columns=3,
            legend style={
                align=center,
                column sep=4ex,
            },
            legend entries={Other, Answer, Question}
        ]
            \addlegendimage{mark=\othermark,solid,color=othercolor}
            \addlegendimage{mark=\answermark,mark size=3pt,solid,color=answercolor}   
            \addlegendimage{mark=\questionmark,solid,color=questioncolor}
        \end{customlegend}
    \end{tikzpicture}            
     
    \caption{Proportion of tokens routed for answer (string match), question, and other tokens by routing component and layer for \slic Large model, averaged over examples in TriviaQA dev set.}
    \label{fig:routing_prop_layer}
\end{figure*}

\begin{figure*}[h!]
    \centering
    \includegraphics[width=0.8\textwidth]{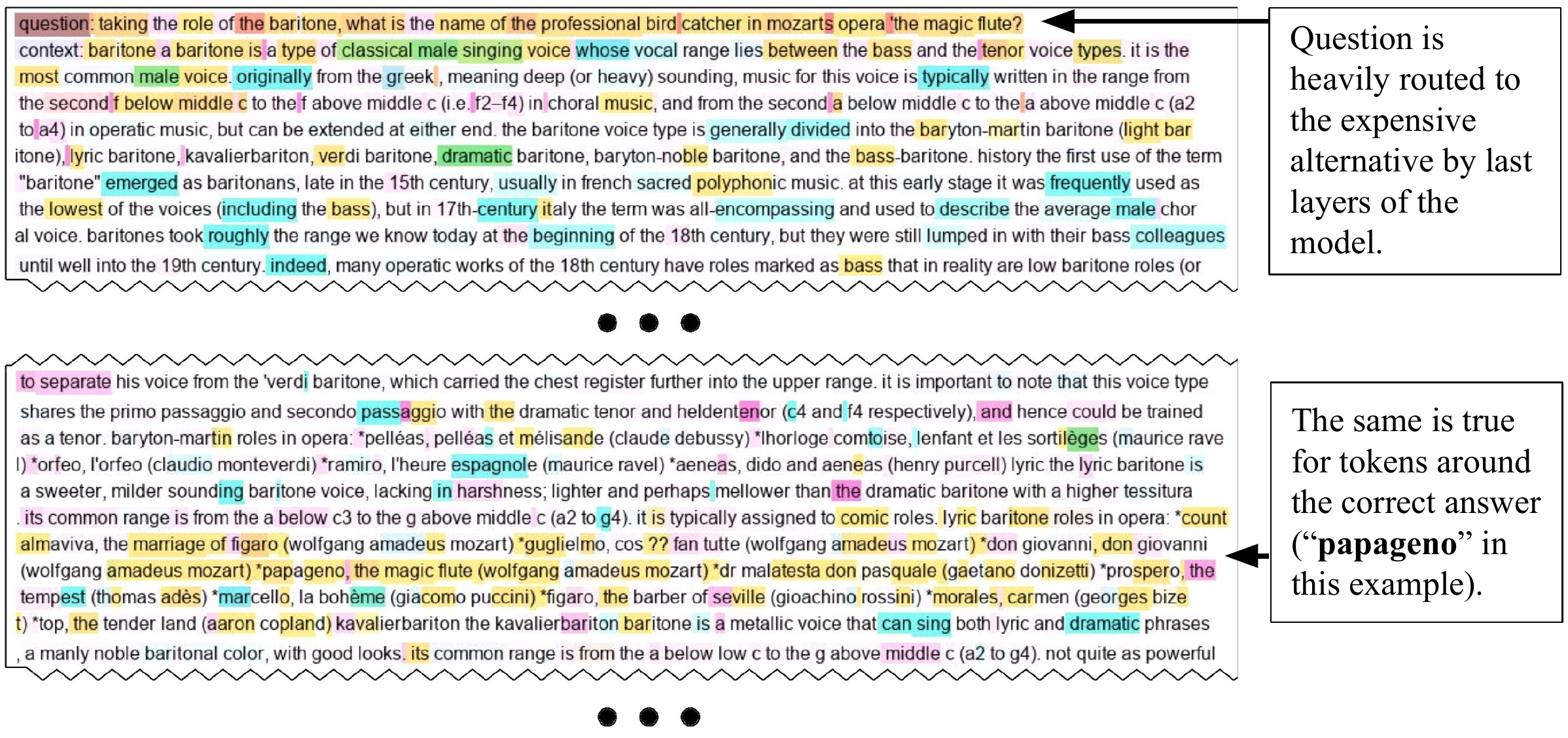}
    \caption{Visualization of token routing weights for some fragments of an example on TriviaQA.}
    \label{fig:routing-viz}
\end{figure*}

\section{Computational Resources}
\label{sec:appendix-compute}

For pre-training we generally used 128 TPUv4 chips for Base and 256 TPUv4 chips for Large and XL. Pre-training took approximately 2.5 days for Base, 3.7 days for Large, and 12.8 days for XL. For fine-tuning we generally used 64, 128, and 256 TPUv4 chips for Base, Large, and XL, respectively, with training time varying with dataset size.

\section{Routing Analysis}
\label{sec:appendix-routing}

In this section we take a closer look at the routing mechanisms in \slic. There are three routing processes in each layer of \slic : (1) Routing of attention keys and values (``KV-routing''), (2) routing of attention queries (``Q-routing'') and (3) routing of MLP tokens (``MLP-routing''). For simplicity, we will say that a token is {\em selected}, when it is routed to the heavy alternative (of either MLP or attention). We are interested in understanding what tokens are selected and whether these mechanisms select similar or different tokens in each layer.

\paragraph{Which tokens are selected} 

We divide input tokens into three categories: (1) question tokens, (2) answer tokens (found via simple normalized string match of the ground truth answer), and (3) other tokens. Figure \ref{fig:routing_prop_layer} shows the proportion of each token type that is routed by a fine-tuned \slic-Large model on the TriviaQA dev set, by layer and routing component. 

Earlier we showed that question and answer tokens are more likely to be selected, but separating routing decisions by layer reveals interesting patterns. At early layers question and answer tokens are only modestly more likely to be selected, with routing probability sharply increasing at later layers and peaking in the last layer. This makes intuitive sense: in early layers the model has not yet had the opportunity to identify which tokens and parts of the document are important. However, the increase is not monotonic and there is strong variation between layers. This variation may imply that different layers focus on different types of tokens, or that some routing components do not successfully learn to identify important tokens.

To gain a better insight into this, Figure \ref{fig:routing-viz} visualizes routing on two sample fragments from a TriviaQA example (notice that, given the large input length used in \slic, we do not show the complete example in the figure). The two fragments shown correspond to the beginning of the example (where the question is located), and the part of the context surrounding the correct answer. We have added a colored background to the figure, where each of the three CMY channels are mapped to the KV-routing weights in different layers of the model. {\em Cyan} corresponds to layer 1, {\em Magenta} to layer 12, and {\em Yellow} to layer 24. As we can see, question and answer are heavily yellow colored, showing those tokens are selected in the last layer.

\paragraph{Correlation between routing processes.} Table \ref{table:routing_correlation} shows the Pearson correlation coefficient between the routing weights of the different routing mechanisms in each layer in a \slic\ {\em Large} model (MLP-routing correlation with KV-routing, MLP-routing with Q-routing, and KV-routing with Q-routing). We show numbers for both the pre-trained checkpoint, as well as a fine-tuned model on TriviaQA. As we can see, the routing of keys/values and routing of queries is highly correlated at all layers except the first two, while the routing of tokens in the MLP has lower correlation to the other two processes. Interestingly correlation between MLP and attention routing increases in the last layers of the model.

\begin{table}[H]
\centering
\footnotesize
\setlength{\tabcolsep}{1pt}
\begin{tabular}{|l|c|c|c||c|c|c|}
\hline
& \multicolumn{3}{c||}{Pre-trained} & \multicolumn{3}{c|}{Fine-tuned} \\ \hline
& {\bf MLP-KV} & {\bf MLP-Q} & {\bf KV-Q} & {\bf MLP-KV} & {\bf MLP-Q} & {\bf KV-Q} \\ \hline
 1 & \correlationBar{-0.06} &	\correlationBar{-0.06} &	\correlationBar{-0.09} &	\correlationBar{-0.06} &	\correlationBar{-0.09} &	\correlationBar{-0.26} \\
 2 & \correlationBar{0.27} &	\correlationBar{0.52} &	\correlationBar{0.04} &	\correlationBar{0.27} &	\correlationBar{0.39} &	\correlationBar{0.02} \\
 3 & \correlationBar{-0.05} &	\correlationBar{-0.03} &	\correlationBar{0.75} &	\correlationBar{0.05} &	\correlationBar{-0.01} &	\correlationBar{0.69} \\
 4 & \correlationBar{0.05} &	\correlationBar{0.09} &	\correlationBar{0.76} &	\correlationBar{0.18} &	\correlationBar{0.14} &	\correlationBar{0.72} \\
 5 & \correlationBar{0.02} &	\correlationBar{-0.01} &	\correlationBar{0.75} &	\correlationBar{0.22} &	\correlationBar{0.26} &	\correlationBar{0.68} \\
 6 & \correlationBar{0.02} &	\correlationBar{-0.01} &	\correlationBar{0.78} &	\correlationBar{0.31} &	\correlationBar{0.33} &	\correlationBar{0.70} \\
 7 & \correlationBar{0.02} &	\correlationBar{0.00} &	\correlationBar{0.73} &	\correlationBar{0.26} &	\correlationBar{0.27} &	\correlationBar{0.70} \\
 8 & \correlationBar{0.00} &	\correlationBar{-0.02} &	\correlationBar{0.44} &	\correlationBar{0.11} &	\correlationBar{-0.07} &	\correlationBar{0.29} \\
 9 & \correlationBar{0.13} &	\correlationBar{0.11} &	\correlationBar{0.74} &	\correlationBar{0.36} &	\correlationBar{0.40} &	\correlationBar{0.70} \\
10 & \correlationBar{-0.06} &	\correlationBar{-0.08} &	\correlationBar{0.08} &	\correlationBar{-0.15} &	\correlationBar{-0.15} &	\correlationBar{0.12} \\
11 & \correlationBar{-0.05} &	\correlationBar{-0.07} &	\correlationBar{0.31} &	\correlationBar{-0.08} &	\correlationBar{-0.03} &	\correlationBar{0.18} \\
12 & \correlationBar{-0.04} &	\correlationBar{-0.08} &	\correlationBar{0.27} &	\correlationBar{0.03} &	\correlationBar{0.00} &	\correlationBar{0.28} \\
13 & \correlationBar{-0.10} &	\correlationBar{-0.09} &	\correlationBar{0.87} &	\correlationBar{-0.13} &	\correlationBar{-0.03} &	\correlationBar{0.72} \\
14 & \correlationBar{-0.04} &	\correlationBar{-0.05} &	\correlationBar{0.76} &	\correlationBar{-0.06} &	\correlationBar{-0.12} &	\correlationBar{0.67} \\
15 & \correlationBar{0.53} &	\correlationBar{0.64} &	\correlationBar{0.69} &	\correlationBar{0.51} &	\correlationBar{0.55} &	\correlationBar{0.67} \\
16 & \correlationBar{0.08} &	\correlationBar{0.12} &	\correlationBar{0.63} &	\correlationBar{0.06} &	\correlationBar{0.57} &	\correlationBar{0.24} \\
17 & \correlationBar{0.28} &	\correlationBar{0.30} &	\correlationBar{0.65} &	\correlationBar{0.27} &	\correlationBar{0.32} &	\correlationBar{0.69} \\
18 & \correlationBar{0.28} &	\correlationBar{0.02} &	\correlationBar{0.84} &	\correlationBar{0.31} &	\correlationBar{0.20} &	\correlationBar{0.76} \\
19 & \correlationBar{0.45} &	\correlationBar{0.77} &	\correlationBar{0.59} &	\correlationBar{0.19} &	\correlationBar{0.38} &	\correlationBar{0.64} \\
20 & \correlationBar{0.30} &	\correlationBar{0.39} &	\correlationBar{0.64} &	\correlationBar{0.38} &	\correlationBar{0.47} &	\correlationBar{0.62} \\
21 &	\correlationBar{0.05} &	\correlationBar{-0.04} &	\correlationBar{0.49} &	\correlationBar{0.18} &	\correlationBar{0.11} &	\correlationBar{0.47} \\
22 & \correlationBar{0.05} &	\correlationBar{0.00} &	\correlationBar{0.69} &	\correlationBar{0.21} &	\correlationBar{0.16} &	\correlationBar{0.68} \\
23 & \correlationBar{0.39} &	\correlationBar{0.33} &	\correlationBar{0.68} &	\correlationBar{0.60} &	\correlationBar{0.79} &	\correlationBar{0.69} \\
24 & \correlationBar{0.43} &	\correlationBar{0.39} &	\correlationBar{0.59} &	\correlationBar{0.57} &	\correlationBar{0.63} &	\correlationBar{0.65} \\
\hline
\end{tabular}
\caption{Pearson correlation coefficient between the routing weights of the different routing mechanisms in each layer in a \slic\ {\em Large} model. We show numbers for both the pre-trained checkpoint, as well as a fine-tuned model on TriviaQA. Blue bars visualize positive correlation, whereas red bars visualize negative correlation.}
\label{table:routing_correlation}
\end{table}

\end{document}